\documentclass[11pt, logo, onecolumn, copyright, colorlinks=true, allcolors=blue]{nvidiatechreport}
\usepackage{graphicx} 
\usepackage{subcaption,ragged2e}
\usepackage[round]{natbib}

\usepackage[normalem]{ulem}
\usepackage[utf8]{inputenc} 
\usepackage[T1]{fontenc}    
\usepackage{url}
\usepackage{tikz}
\usepackage{enumitem}
\usetikzlibrary{positioning}
\usetikzlibrary{calc}
\usepackage{array}
\usepackage{booktabs}
\usepackage{wrapfig}
\usepackage{caption} 
\usepackage{listings}
\usepackage{adjustbox}
\usepackage{footnote}
\usepackage{multirow}
\usepackage{amsmath}
\usepackage{amsfonts}
\usepackage{breakcites}
\usepackage{blindtext}
\usepackage{svg}
\usepackage[most]{tcolorbox}
\usepackage{tabularx}
\usepackage{graphicx} 
\newcommand{\newparagraph}[1]{\noindent\textbf{#1\hspace{0.5em}}}
\newcommand{\mathdataset}[0]{Nemotron-CC-Math}

\newenvironment{simplechar}{%
   \catcode`\$=12
   \catcode`\&=12
   \catcode`\#=12
   \catcode`\^=12
   \catcode`\_=12
   \catcode`\~=12
   \catcode`\%=12
}{}

\newcommand{\ourmodelfull}{NVIDIA Nemotron Nano 2\xspace}
\newcommand{\ourmodel}{Nemotron Nano 2\xspace}
\newcommand{\ourbasemodel}{Nemotron-Nano-12B-v2-Base\xspace}
\newcommand{\ourprunedbasemodel}{Nemotron-Nano-9B-v2-Base\xspace}
\newcommand{\ourfinalmodel}{Nemotron-Nano-9B-v2\xspace}

\title{\ourmodelfull: An Accurate and Efficient Hybrid Mamba-Transformer Reasoning Model}
\author{\large NVIDIA}
\date{}

\begin{document}

\begin{abstract}
\large \textbf{Abstract.}
\normalsize
We introduce \ourfinalmodel, a hybrid Mamba-Transformer language model designed to increase throughput for reasoning workloads while achieving state-of-the-art accuracy compared to similarly-sized models. \ourfinalmodel builds on the Nemotron-H architecture, in which the majority of the self-attention layers in the common Transformer architecture are replaced with Mamba-2 layers, to achieve improved inference speed when generating the long thinking traces needed for reasoning. We create \ourfinalmodel by first pre-training a 12-billion-parameter model (\ourbasemodel) on 20 trillion tokens using an FP8 training recipe. After aligning \ourbasemodel, we employ the Minitron strategy to compress and distill the model with the goal of enabling inference on up to 128k tokens on a single NVIDIA A10G GPU (22GiB of memory, \texttt{bfloat16} precision). Compared to existing similarly-sized models (e.g., Qwen3-8B), we show that \ourfinalmodel achieves on-par or better accuracy on reasoning benchmarks while achieving up to 6$\times$ higher inference throughput in reasoning settings like 8k input and 16k output tokens (Figure~\ref{fig:intro}). We are releasing \ourfinalmodel, \ourbasemodel, and \ourprunedbasemodel checkpoints along with the majority of our pre- and post-training datasets on Hugging Face.
\end{abstract}

\maketitle

\section{Introduction}
\label{sec:intro}

\begin{figure}[htbp]
    \centering
    \includegraphics[width=\linewidth]{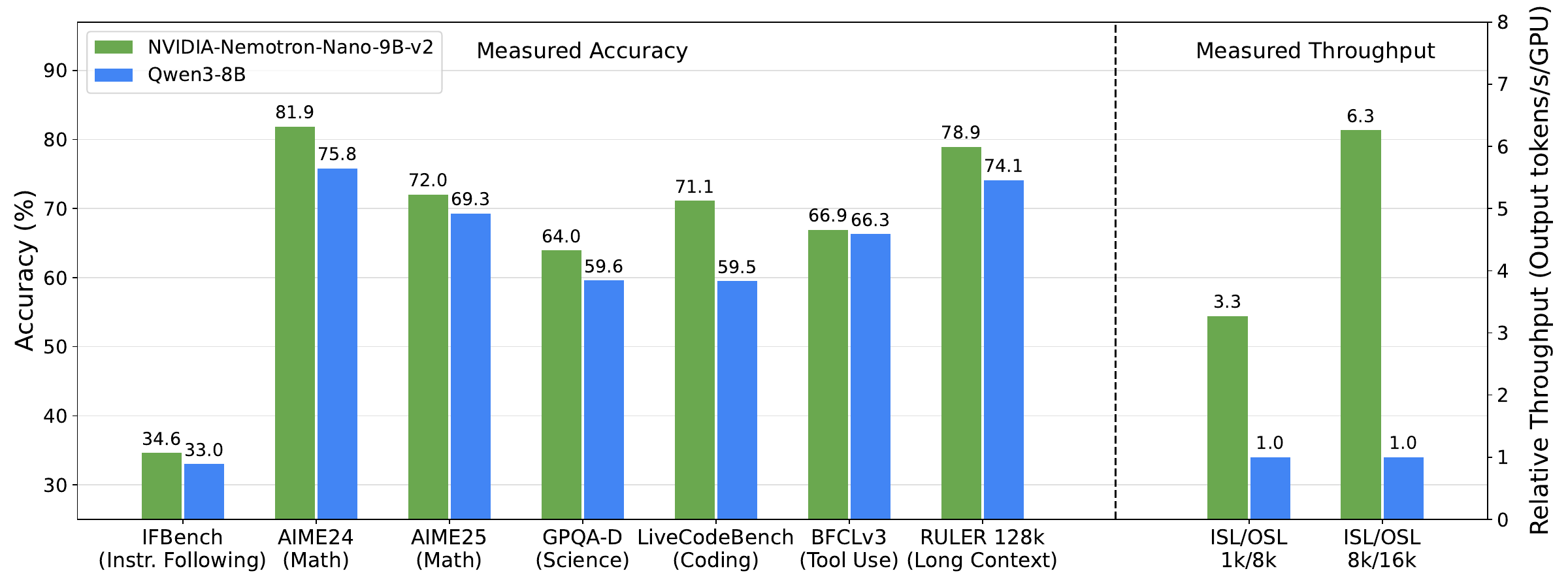}
    \caption{Comparison of \ourmodel and Qwen3-8B in terms of accuracy and throughput. \ourmodel achieves comparable or better accuracies on complex reasoning benchmarks, while achieving up to 6.3$\times$ higher throughput for such workloads. We abbreviate input sequence length to ISL and output sequence length to OSL and measure throughput on a single A10G GPU in \texttt{bfloat16}.}
    \label{fig:intro}
\end{figure}

We introduce \ourmodelfull, a hybrid Mamba-Transformer reasoning model~\citep{waleffe2024empiricalstudymambabasedlanguage,lieber2024jambahybridtransformermambalanguage, gemmateam2025gemma3technicalreport,nvidia2025nemotronhfamilyaccurateefficient} that achieves on-par or better benchmark accuracies at 3$\times$--6$\times$ higher throughput than Qwen3-8B~\citep{yang2025qwen3technicalreport} for generation-heavy scenarios like 1k input / 8k output or 8k input / 16k output tokens (Figure~\ref{fig:intro}). \ourmodel builds on the architecture of Nemotron-H~\citep{nvidia2025nemotronhfamilyaccurateefficient}, but utilizes key new datasets and recipes for pre-training, alignment, pruning and distillation. We share these recipes, the checkpoints, as well as the majority of the pre- and post-training datasets.

The initial base model, \ourbasemodel, was pre-trained using FP8 precision~(\S\ref{sec:fp8}) over 20 trillion tokens using a Warmup-Stable-Decay~\citep{hu2024minicpm} learning rate schedule~(\S\ref{sec:pre-train-hyperparams}). It then underwent a continuous pre-training long-context extension phase to become 128k-capable without degrading other benchmarks~(\S\ref{sec:pre-train-long-context}). Overall, new and improved datasets led to significant accuracy improvements over Nemotron-H-8B on math, multilingual, MMLU-Pro and other benchmarks~(\S\ref{sec:pretrain_data}).

\ourmodel was then post-trained through a combination of Supervised Fine-Tuning (SFT), Group Relative Policy Optimization (GRPO)~\citep{Shao2024DeepSeekMath}, Direct Preference Optimization (DPO)~\citep{Rafailov2023DPO}, and Reinforcement Learning from Human Feedback (RLHF)~\citep{Ouyang2022InstructGPT,Christiano2017DeepRLHF}. We applied multiple SFT stages across various domains, followed by targeted SFT on key areas such as tool use, long-context performance, and truncated (budgeted) training. 
GRPO and RLHF sharpened instruction-following and conversational ability, while additional DPO stages further strengthened tool use. 
Overall, post-training was performed on roughly $90$ billion tokens, the majority in single-turn prompt–response format with reasoning traces. 
About $5\%$ of the data contained deliberately truncated reasoning traces, enabling fine-grained thinking budget control at inference time (\S\ref{sec:budgetcontrol}).

Finally, both the base model and aligned model were compressed so as to enable inference over context lengths of 128k tokens on a single NVIDIA A10G GPU (22 GiB of memory, \texttt{bfloat16} precision). This was done by extending a compression strategy based on Minitron~\citep{muralidharan2024compactlanguagemodelspruning,sreenivas2024llmpruningdistillationpractice,taghibakhshi2025efficient} to compress reasoning models subject to constraints.

We are releasing the following models on Hugging Face:
\begin{itemize}
    \item \href{https://huggingface.co/nvidia/NVIDIA-Nemotron-Nano-9B-v2}{\textbf{NVIDIA-Nemotron-Nano-9B-v2}}:
       the aligned and pruned reasoning model,
    \item \href{https://huggingface.co/nvidia/NVIDIA-Nemotron-Nano-9B-v2-Base}{\textbf{NVIDIA-Nemotron-Nano-9B-v2-Base}}:
       a pruned base model,
    \item \href{https://huggingface.co/nvidia/NVIDIA-Nemotron-Nano-12B-v2-Base}{\textbf{NVIDIA-Nemotron-Nano-12B-v2-Base}}:
      the base model before alignment or pruning.
\end{itemize}
Additionally, we are releasing the majority of our pre-training dataset in the \href{https://huggingface.co/collections/nvidia/nemotron-pre-training-dataset-689d9de36f84279d83786b35}{\textbf{Nemotron-Pre-Training-Dataset-v1}} collection of more than 6 trillion tokens:
\begin{itemize}
    \item \href{https://huggingface.co/datasets/nvidia/Nemotron-CC-v2}{\textbf{Nemotron-CC-v2}}: Follow-up to Nemotron-CC~\citep{su2024nemotroncctransformingcommoncrawl} with eight additional Common Crawl snapshots (2024–2025), synthetic rephrasing, deduplication, and synthetic Q\&A data translated into 15 languages.
    \item \href{https://huggingface.co/datasets/nvidia/Nemotron-CC-Math-v1}{\textbf{Nemotron-CC-Math-v1}}: 133B-token math dataset from Common Crawl using Lynx + LLM pipeline~\citep{karimi2025nemotronccmath}. Preserves equations, standardizes to LaTeX, outperforms previous math datasets on benchmarks.
    \item \href{https://huggingface.co/datasets/nvidia/Nemotron-Pretraining-Code-v1}{\textbf{Nemotron-Pretraining-Code-v1}}: Curated GitHub code references with multi-stage filtering, deduplication, and quality filters. Includes code Q\&A data in 11 programming languages.
    \item \href{https://huggingface.co/datasets/nvidia/Nemotron-Pretraining-SFT-v1}{\textbf{Nemotron-Pretraining-SFT-v1}}: Synthetic SFT-style dataset covering STEM, multilingual, academic, and reasoning domains.
\end{itemize}
Finally, we are releasing an updated post-training dataset:
\begin{itemize}
    \item \href{https://huggingface.co/datasets/nvidia/Nemotron-Post-Training-Dataset-v2}{\textbf{Nemotron-Post-Training-Dataset-v2}}: Adds to NVIDIA’s post-training dataset releases with an extension of SFT and RL data into five target languages: Spanish, French, German, Italian and Japanese. The data supports improvements of math, code, general reasoning, and instruction following capabilities.
\end{itemize}

The rest of this technical report is organized as follows:
In \S\ref{sec:pretraining}, we discuss the \ourmodel model architecture, pre-training process, and base model evaluation results.
In \S\ref{sec:alignment}, we discuss the alignment process.
In \S\ref{sec:compression}, we describe the pruning and distillation methods used for model compression.

\section{Pretraining}
\label{sec:pretraining}

In this section, we discuss the architecture and pretraining of the \ourbasemodel model. We also compare this model against other state-of-the-art models in terms of accuracy on popular benchmarks.

\begin{figure}[!t]
\centering
\includegraphics[width=0.9\linewidth]{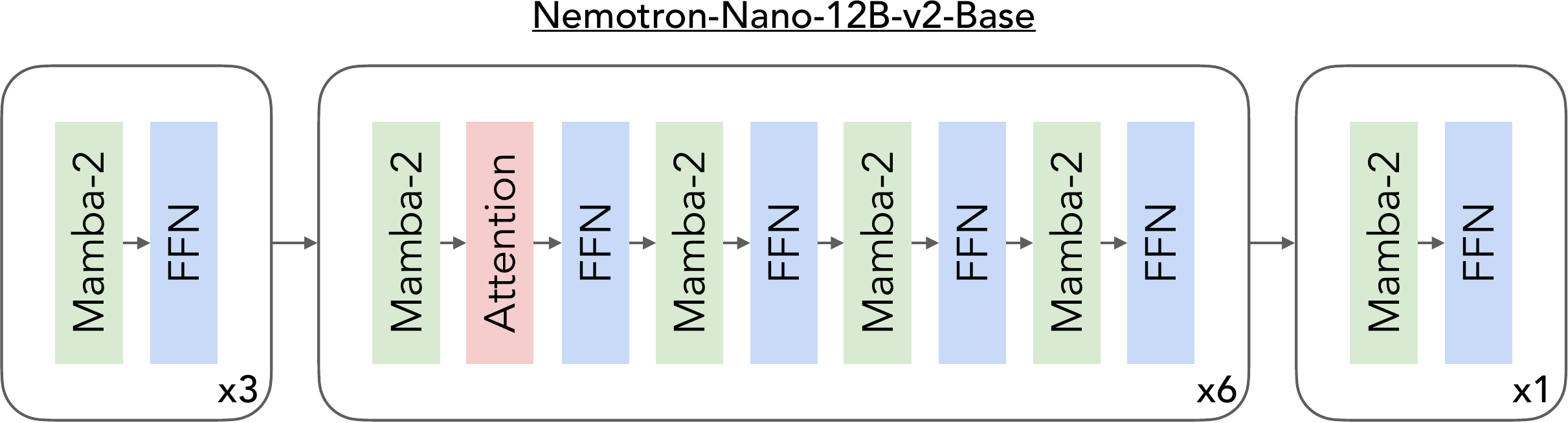}
\caption{\ourbasemodel layer pattern. As in Nemotron-H models, roughly 8\% of the total layers in the model are self-attention layers which are evenly dispersed throughout the model.}
\label{fig:base-12b-layer-pattern}
\end{figure}

\begin{table}[!t]\scriptsize
\centering
\begin{tabular}{lcccccccc}
\toprule
        \textbf{Model} &
        \textbf{\begin{tabular}[c]{@{}c@{}}{Number of}\\ layers\end{tabular}} &
        \textbf{\begin{tabular}[c]{@{}c@{}}{Model}\\ dimension\end{tabular}} &
        \textbf{\begin{tabular}[c]{@{}c@{}}FFN \\ dimension\end{tabular}} &
        \textbf{\begin{tabular}[c]{@{}c@{}}Q\\ heads\end{tabular}} &
        \textbf{\begin{tabular}[c]{@{}c@{}}KV \\ heads\end{tabular}} &
        \textbf{\begin{tabular}[c]{@{}c@{}}State \\ dimension\end{tabular}} & 
        \textbf{\begin{tabular}[c]{@{}c@{}}Mamba \\ groups\end{tabular}} \\ \toprule

\ourbasemodel & 62 & 5120 & 20480 & 40 & 8 & 128 & 8 \\
\bottomrule
\end{tabular}
\caption{Summary of \ourbasemodel architecture.}
\label{tab:base-12b-arch}
\end{table}

\subsection{Model Architecture}
As in Nemotron-H~\citep{nvidia2025nemotronhfamilyaccurateefficient}, \ourbasemodel 
consists of a mixture of Mamba-2~\citep{dao2024transformersssmsgeneralizedmodels}, self-attention, and FFN layers. The layer pattern and key architecture details are summarized in Figure~\ref{fig:base-12b-layer-pattern} and Table~\ref{tab:base-12b-arch}. Concretely, we use 62 layers, with 6 of them being self-attention layers, 28 being FFN, and 28 being Mamba-2 layers. We use a hidden dimension of 5120, FFN hidden dimension of 20480, and Grouped-Query Attention~\citep{ainslie2023gqatraininggeneralizedmultiquery} with 40 query heads and 8 key-value heads. For Mamba-2 layers, we use 8 groups, a state dimension of 128, a head dimension of 64, an expansion factor of 2, and a window size for convolution of 4. For FFN layers, we use squared ReLU~\citep{so2022primersearchingefficienttransformers} activation. Again as in Nemotron-H, we do not use any position embeddings and use RMSNorm~\citep{zhang2019rootmeansquarelayer}, separate embedding and output layer weights, no dropout, and we do not use bias weights for linear layers.

\subsection{Pre-Training Data}
\label{sec:pretrain_data}
\ourbasemodel was pre-trained on a large corpus of high-quality curated and synthetically-generated data.

\subsubsection{Curated Data}

We have separate data curation pipelines for the following broad data categories: general web crawl data (English and multilingual), math data, and code data. We discuss each in turn next.

\newparagraph{English web crawl data.} We used the Nemotron-CC dataset~\citep{su2024nemotroncctransformingcommoncrawl}, but updated to include eight more recent Common Crawl snapshots (CC-MAIN-2024-33 through CC-MAIN-2025-13) using the same pipeline. For synthetic rephrasing, we mostly switched to Qwen3-30B-A3B (from Mistral Nemo 12B). Additionally, we used data from CC-NEWS through April 23, 2025, to help improve the knowledge cutoff of the model. The CC-NEWS data was filtered for English and globally fuzzily de-duplicated; no other filtering was used.

\newparagraph{Multilingual data.} We extracted data for fifteen languages from the following three Common Crawl snapshots: CC-MAIN-2024-51, CC-MAIN-2025-08, and CC-MAIN-2025-18. The fifteen languages included were Arabic, Chinese, Danish, Dutch, French, German, Italian, Japanese, Korean, Polish, Portuguese, Russian, Spanish, Swedish, and Thai. As we did not have reliable multilingual model-based quality classifiers available, we just applied heuristic filtering instead. This was done in a similar manner to the filtering of low-quality English data in the Nemotron-CC pipeline, except that we had to selectively disable some heuristic filters that had very high false positive rates for some languages. De-duplication was done in the same way as for Nemotron-CC. Additionally, we used data from Wikipedia and FineWeb-2~\citep{penedo2025fineweb2pipelinescale} for these fifteen languages.

\newparagraph{Math data.} Mathematical content on the web is expressed in a wide range of formats, including inline and block \LaTeX, MathML, Unicode symbols, and custom renderers such as MathJax or KaTeX. We conducted a detailed analysis of prior math-specific extraction pipelines—including OpenWebMath~\citep{paster2023openwebmath}, MegaMath~\citep{zhou2025megamath}, jusText~\citep{justext}, Trafilatura~\citep{barbaresi-2021-trafilatura}, and Resiliparse~\citep{bevendorff:2018}—and found that none could reliably preserve mathematical expressions or code structure. These tools frequently discard or distort equations and flatten code formatting, severely limiting the utility of the extracted content for pretraining.

To address this, we built a new pipeline specifically designed for high-fidelity mathematical extraction from Common Crawl. We first aggregated a comprehensive list of math-related URLs from prior datasets (e.g., InfiMM-WebMath~\citep{han2024infimm}, OpenWebMath~\citep{paster2023openwebmath}, FineMath~\citep{allal2025smollm2smolgoesbig}, and MegaMath~\citep{zhou2025megamath}), then re-fetched their raw HTML documents from 98 Common Crawl snapshots (2014–2024). Each page was rendered using the \texttt{lynx} text-based browser to preserve layout and math structure. We then applied Phi-4~\citep{abdin2024phi}(14B-parameters) to remove boilerplate, standardize notation into \LaTeX, and correct inconsistencies. A FineMath classifier~\citep{allal2025smollm2smolgoesbig} was used to retain high-quality documents, followed by fuzzy deduplication via MinHash-based~\citep{broder2000identifying} Locality Sensitive Hashing (LSH)~\citep{indyk1998approximate} via the NeMo-Curator framework.\footnote{\url{https://github.com/NVIDIA-NeMo/Curator}} We finally decontaminated the dataset using LLM Decontaminator~\citep{yang2023rethinking}.

This process resulted in a 133B-token corpus, \mathdataset-3+, and a higher-quality 52B-token subset, \mathdataset-4+, containing only the top-scoring samples. When used for pretraining, this dataset yields substantial improvements across math (MATH-500), code (HumanEval+, MBPP+, MBPP), and general-domain evaluations (MMLU, MMLU-STEM, MMLU-Pro), surpassing all existing open math datasets. For full details, see~\citet{karimi2025nemotronccmath}.

\newparagraph{Code data.} In line with previous models in the Nemotron family \citep{nvidia2025nemotronhfamilyaccurateefficient,nvidia2024nemotron4340btechnicalreport,parmar2024nemotron415btechnicalreport}, we pretrained \ourbasemodel with large-scale raw source code. All source code used to train this model originated from GitHub and went through a multi-stage processing pipeline to arrive at the final source code training data. We performed license-based removal with a license detection pipeline similar to that used by the BigCode project \citep{lozhkov2024starcoder2stackv2}, but with fewer accepted licenses (see Appendix~\ref{appendix:acceptable-licenses} for additional details). De-duplication is especially important for source code, where many files can be found exactly duplicated across numerous repositories. Consequently we performed both exact (via hashing) and fuzzy deduplication (using MinHash LSH). In order to build a better understanding of each file in our dataset, we annotated all files with a variety of measures and then performed filtering using these annotations. We found the heuristic filters from OpenCoder~\citep{huang2025opencoderopencookbooktoptier} to be effective and leveraged them to filter files that are less valuable or even detrimental for LLM pretraining.

\subsubsection{Synthetically-Generated Data}
\label{section:synthetic}

\newparagraph{STEM data.} We generated synthetic data for STEM subjects, including Astronomy, Biology, Chemistry, Math, and Physics using 88.6k questions collected from multiple sources as the seed data.
In addition to the widely used GSM8K, MATH, and AOPS training sets, we collected more diverse questions from Stemez\footnote{\url{https://www.stemez.com/}} and textbooks with permissive licenses from OpenStax\footnote{\url{https://openstax.org}} and Open Textbook Library.\footnote{\url{https://open.umn.edu/opentextbooks/}} We used Qwen2.5-VL-72B-Instruct~\citep{bai2025qwen25vltechnicalreport} to extract questions from the exercise sections in the textbooks with additional instructions such as dropping question numbering, ignoring questions that require image interpretation, and formatting equations using LaTeX. We manually curated the extracted questions to fix occasional OCR errors and removed non-self-contained questions (e.g., a question that refers to an example in the same chapter).

To expand both the quantity and diversity of questions, we conducted three iterations of question generation using four models (i.e., Qwen3-30B-A3B and Qwen3-235B-A22B~\citep{yang2025qwen3technicalreport}, both with thinking mode enabled, Deepseek-R1~\citep{deepseekai2025deepseekr1incentivizingreasoningcapability}, and Deepseek V3~\citep{deepseekai2025deepseekv3technicalreport}) and three prompts:
\begin{enumerate}
\item \textbf{Similar question:} Create a new question that explores similar concepts but offers a fresh challenge.
\item \textbf{Harder question:} Create a new question that requires more logical steps or involves more advanced concepts.
\item \textbf{Varied question:} Create a new question that differs in type from the original question. We instructed the model to avoid superficial or trivial modifications and think through the solution when creating a new question.
\end{enumerate}
We filtered out duplicates and highly-similar questions using fuzzy de-duplication and generated solutions to the remaining questions with the models used in the question generation step. We converted a subset of examples to multiple-choice questions in MMLU or MMLU-Pro style. We constructed a few thousand few-shot examples by concatenating random synthetic samples.

\newparagraph{Math data.} We also revisited and regenerated the Nemotron-MIND dataset~\citep{akter2024mindmathinformedsynthetic}, a math-informed synthetic pretraining corpus originally built on OpenWebMath. In our updated version, we regenerated the MIND dataset using \mathdataset-4+, our highest-quality math subset comprising 52B tokens—as the source corpus. Following the original methodology, we applied seven prompt templates (e.g., Teacher–Student, Debate, Interview, etc) to generate structured mathematical dialogues using the Phi-4 model. Unlike the original MIND, which relied on 14.7B tokens of lower-fidelity data, our version leverages significantly higher-quality input and processes it with a chunk size of 5K tokens. This regeneration produced a 73B-token synthetic dataset and led to consistent improvements across math reasoning and general knowledge (MMLU, MMLU-Pro. MMLU-Stem) benchmarks compared to the original MIND version, highlighting the critical role of input data quality. Full details and results are available in~\citet{karimi2025nemotronccmath}.

\newparagraph{Multilingual data.} We generated multilingual diverse question and answer data (Diverse QA)~\citep{su2024nemotroncctransformingcommoncrawl} from two sources:
\begin{enumerate}
\item We translated the English Diverse QA data to fifteen languages (see Multilingual data) using Qwen3-30B-A3B~\citep{yang2025qwen3technicalreport}.
\item We generated synthetic data from Wikipedia articles in these languages using the Diverse QA prompt and instructed the model to write all questions and answers in the target language.
\end{enumerate}
In addition, we translated a subset of our GSM8K augmentation data (see STEM data) into these languages using Qwen3-30B-A3B. We post-processed each translated solution by appending a concluding sentence meaning ``\textit{The answer is ...}'' (e.g., ``\textit{La respuesta es ...}'' in Spanish, ``\textit{Die Antwort lautet ...}'' in German), where the final numerical answer is extracted from the original English solution.

\newparagraph{Code data.} We generated question-answer (QA) data at scale for 11 different programming languages by prompting an LLM to generate questions based on short snippets from our curated source code, asking the model to solve the generated question, and then performing post hoc filtering of the generated QA pairs based on heuristics as appropriate (e.g., Python AST parsing). This technique results in diverse synthetic data targeted at problem solving containing both natural language and source code. Further details are covered in the Nemotron-H technical report \citep{nvidia2025nemotronhfamilyaccurateefficient}, where we first leveraged this type of synthetic code data in pretraining.

\newparagraph{Academic data.}
In the pretraining set for the Nemotron-H \citep{nvidia2025nemotronhfamilyaccurateefficient} series of models, we assigned attribute labels for educational quality, educational difficulty, and educational subject to all documents coming from academic data, which encompasses textbooks and academic papers. As content of higher educational difficulty in technical domains still proves challenging for models, we prioritized increasing model comprehension of such information in our current pretraining set via the generation of question-answer (QA) pairs as such data has been shown to enhance knowledge storage and extraction within language models \citep{allenzhu2024physicslanguagemodels31}.   

To do so, we first gathered all documents with educational difficulty at the undergraduate and graduate levels in the following technical subject areas: math, chemistry, biology, physics, and medicine. Using this subset of documents, we aim to find the most relevant pieces of texts that could be utilized as seed contexts for our generation of QA pairs. We chunk each document into snippets of 512 token lengths, embed them with the e5-large model \citep{wang2024textembeddingsweaklysupervisedcontrastive}, and store them within a Milvus vector database that enables approximate nearest neighbor search. We then curate documents from a set of complex subject areas (e.g. Mathematics: Real Analysis, Biology: Genetics, Statistics: Information Theory), and query the Milvus database for the 250 nearest neighbor text snippets to each query document. The returned snippets function as our seed contexts that we then pass into a Qwen-2.5 72B instruct model \citep{qwen2025qwen25technicalreport} to generate multiple choice and free response style QA pairs based on the information contained in the snippet. With each QA pair, a justification for the answer is additionally generated.

\newparagraph{SFT-style data.} 
Using SFT-style data in the later stages of pretraining has shown to be helpful to foster more comprehensive model learning~\citep{hu2024minicpm}.

Therefore, we synthesized and included different SFT-style data covering several domains: 1) code SFT data which is mainly focused on solving code problems; 2) math SFT data that is mostly focused on reasoning; 3) MMLU-style SFT data which contains different question and answer examples covering different knowledge topics; and 4) general instruction following SFT data.

We ensure that the SFT-style data covers diverse topics with different difficulty levels for each of the above mentioned domains. Detailed synthesis methods and pipelines for the above mentioned SFT data can be found in prior work~\citep{toshniwal2024openmath2, moshkov2025aimo2,bercovich2025llamanemotronefficientreasoningmodels, bercovich2025llama,ahmad2025opencodereasoning,ahmad2025opencodeinstruct,majumdar2024genetic}.

\newparagraph{Fundamental reasoning SFT-style data.} While the above mentioned SFT-style data help enhance an LLM's ability to answer questions in code, math and general language understanding benchmarks, they do not help improve the model's ability in deeper reasoning tasks to discern the correct answer among a larger pool of potential distractors. We propose to mitigate that by synthesizing SFT-style data focused on analytical reasoning, logical reasoning, and reading comprehension. 

Specifically, we collected existing datasets including 1) the Law School Admission Test (LSAT) dataset from \citet{wang2022lsat, zhong2022analytical} which encompasses three tasks: logical reasoning,
reading comprehension, and analytical reasoning, 2) the repurposed LogiQA dataset by~\cite{liu2020logiqa} which contains various types of logical reasoning questions collected from the National Civil Servants Examination of China, and 3) the AQuA-RAT dataset which emphasizes algebraic word problems by~\cite{ling2017program}. We then prompted DeepSeek-V3~\citep{deepseekai2025deepseekv3technicalreport} and Qwen3-30B-A3B~\citep{yang2025qwen3technicalreport} respectively to synthesize more similar questions with corresponding options. For each question we generated, we prompted DeepSeek-V3 again to generate the chain-of-thought (CoT) process with the final solution. At the post-processing stage, we apply majority voting to keep only the samples that have the most voted solutions. Overall, we generated 4B tokens from DeepSeek-V3 and 4.2B tokens from Qwen3-30B models.

\subsection{Data Mixture and Ordering}
\label{section:blend}

Our data mixture consists of thirteen data categories. The largest is web crawl data, which we subdivided into four categories based on the Nemotron-CC quality classification~\citep{su2024nemotroncctransformingcommoncrawl}: crawl-medium, crawl-medium-high, crawl-high, syn-crawl-high denoting medium, medium-high, high and synthetic quality crawl data, respectively. 
Apart from these, our data mixture has additional categories such as math, wikipedia, code,
academic data, crawl++, multilingual, and synthetic SFT-style data which is further categorized as general-sft, stem-sft and code-sft. 
Crawl++ consists of web-crawl derivatives like OpenWebText, BigScience and Reddit. 
Our multilingual data has fifteen languages:
Arabic, Danish, German, Spanish, French, Italian, Portuguese, Dutch, Polish, Swedish, Thai, Chinese, Japanese, Korean, and Russian. 
We design the data mixtures to give similar weight to data sources that have similar quality.
Data sources of higher quality are weighed higher than data sources of lower quality. 
We provide detailed explanation on quality estimation of datasets and the blend creation process in \cite{feng2024maximizedataspotentialenhancing} and \cite{nvidia2025nemotronhfamilyaccurateefficient}.

\begin{figure}[htbp]
    \centering
    \begin{subfigure}{0.7\textwidth}
        \centering
        \includegraphics[width=\linewidth]{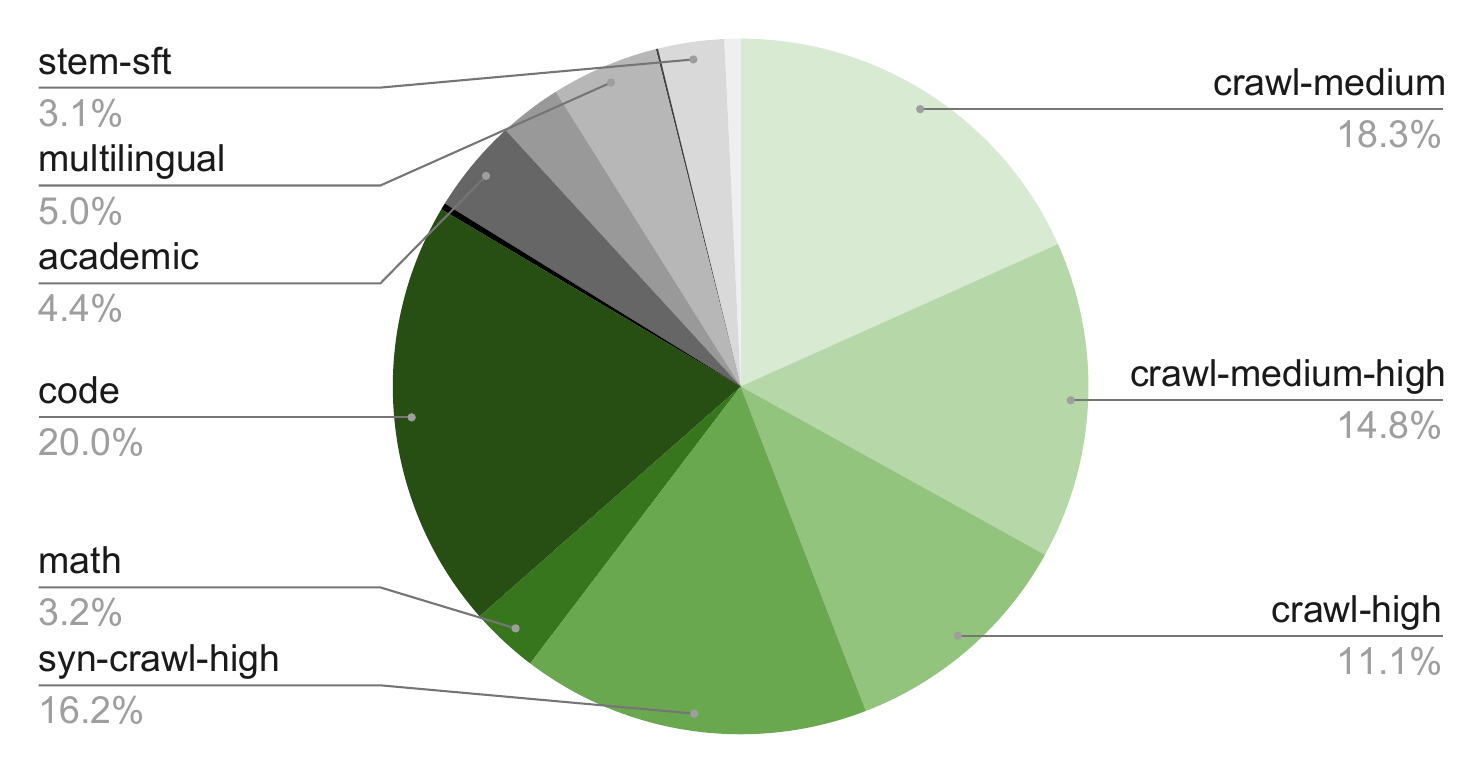}
        \caption{Data mixture of Phase 1.}
        \label{fig:phase1-blend}
    \end{subfigure}
\bigskip
    \begin{subfigure}{0.7\textwidth}
        \centering
        \includegraphics[width=\linewidth]{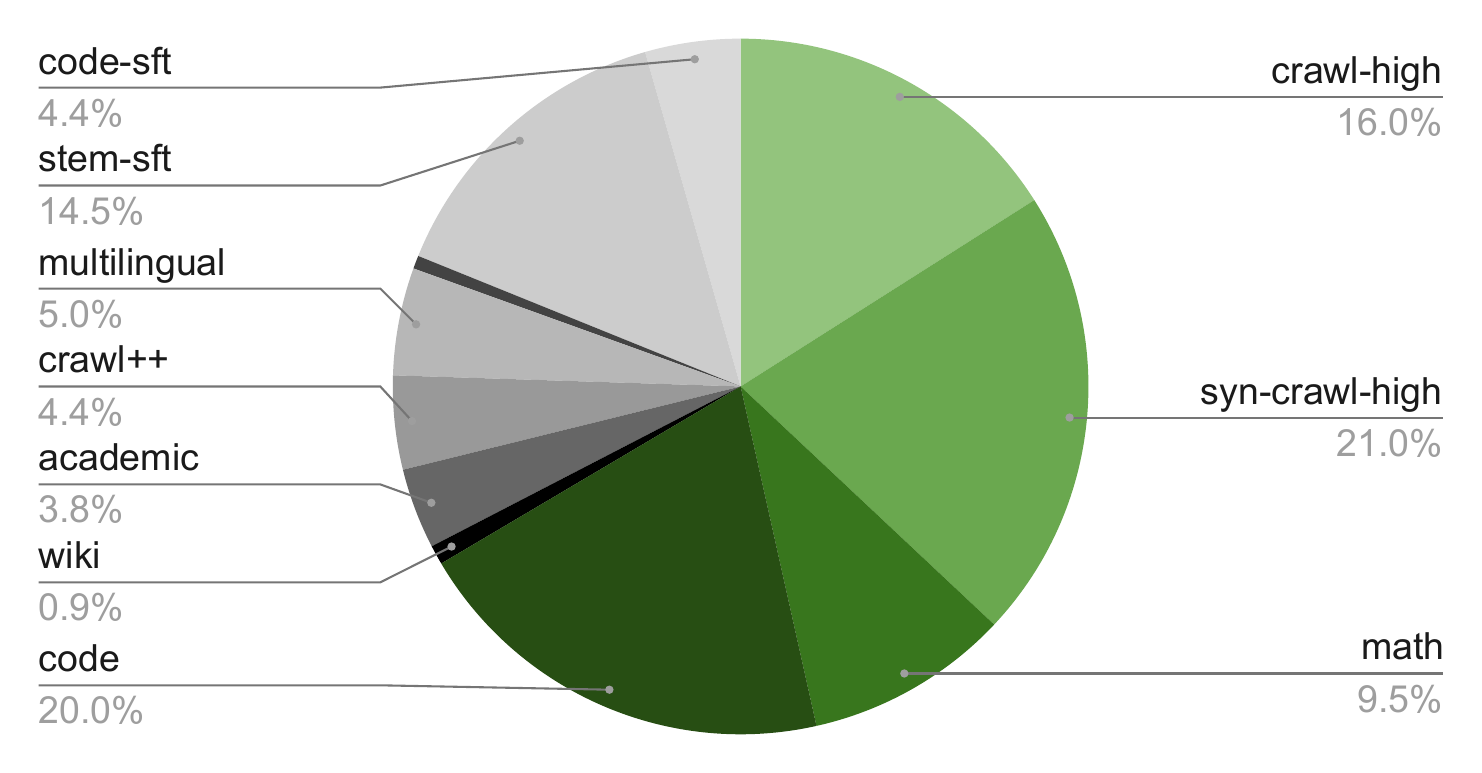}
        \caption{Data mixture of Phase 2.}
        \label{fig:phase2-blend}
    \end{subfigure}
\bigskip
  \begin{subfigure}{0.7\linewidth}
    \centering
    \includegraphics[width=\linewidth]{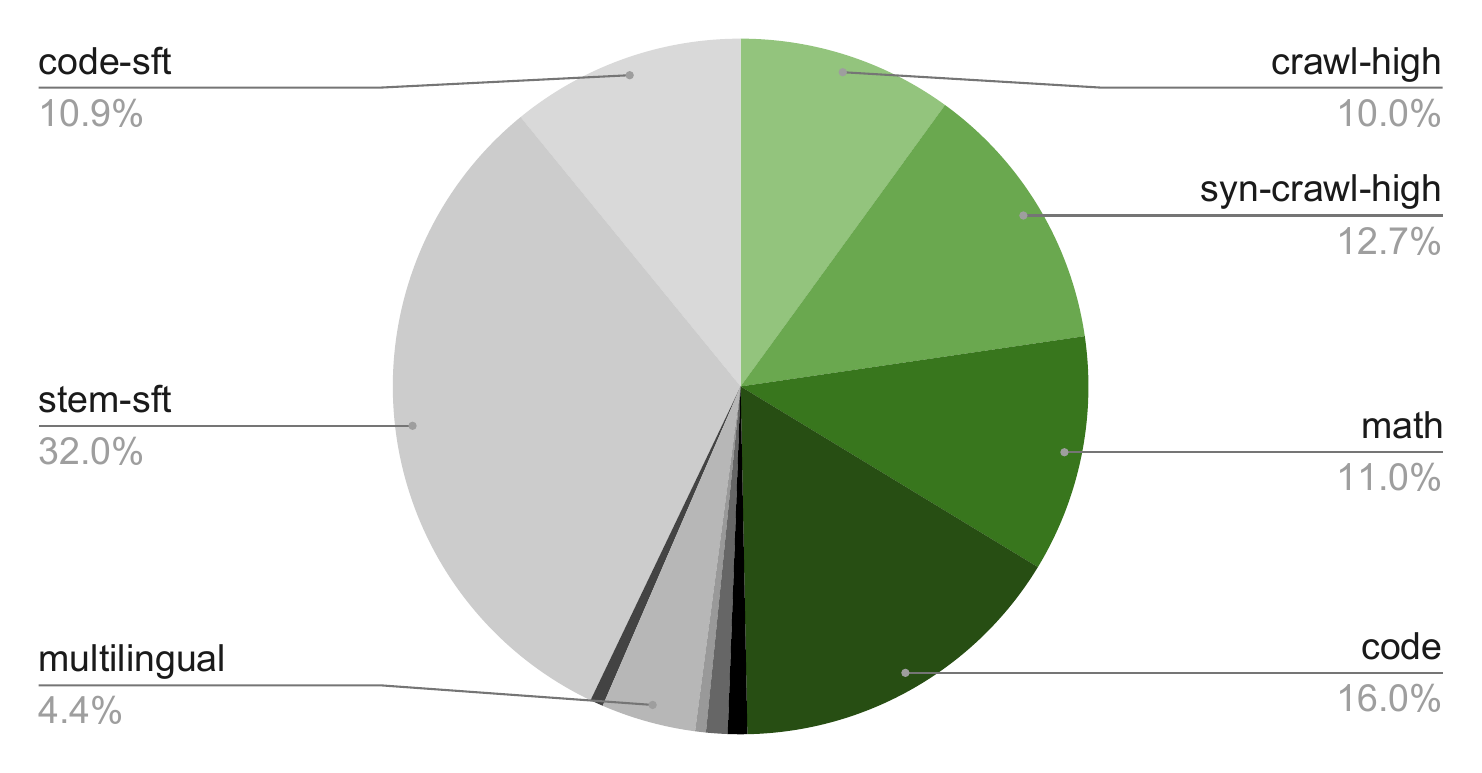}
    \caption{Data mixture of Phase 3.}
    \label{fig:phase3-blend}
  \end{subfigure}
    \caption{Data mixtures for each phase of pre-training.}
    \label{fig:phase-blends}
\end{figure}

We used a curriculum based on three phases of data-blending approach to pre-train \ourbasemodel. 
In the first phase, we used a data mixture that promotes diversity in data; in the second and third phases, we primarily used high-quality datasets (e.g., Wikipedia). 
We switched to the second phase at the 60\% point of training, and to the third phase at the 90\% point of training. 
The data mixtures used in each phase are shown in Figure~\ref{fig:phase-blends}. 

\subsubsection{Multilingual Data Ablation Study}

In Section~\ref{sec:pretrain_data}, we mentioned several large categories of multilingual data, both curated and synthetic:
\begin{enumerate}
    \item \textbf{Common Crawl:} Extracted from recent Common Crawl snapshots using our own pipeline.
    \item \textbf{FineWeb-2}~\citep{penedo2025fineweb2pipelinescale}.
    \item \textbf{DiverseQA-wiki:} Generated from multilingual Wikipedia articles using a translated Diverse QA prompt.
    \item \textbf{DiverseQA-crawl:} Translated from English Diverse QA data.
\end{enumerate}
In order to decide the proper data mixture among these different multilingual data sources, we first conducted ablation experiments to compare the four multilingual data's downstream tasks' performance.

Specifically, we took a 1B model checkpoint that had been trained for 350B tokens, and continuous pretrained it for another 100B tokens. We assigned 50\% of the continuous pretraining data to multilingual data, and the remaining 50\% use our default pretraining data mixture. We evaluated each model's performance using the Global-MMLU benchmark~\citep{singh2024global}; the results are shown in Table~\ref{table:ablation_multi}. Our curated Common Crawl-based multilingual data performed slightly better than the Fineweb2-based multilingual data, while the synthesized multilingual QA pairs performed much better than the curated multilingual web crawl data. The diverse pairs translated from  English Common Crawl achieved the highest average score over the 8 languages we evaluated on. Therefore, we assigned a much higher weight to the DiverseQA-crawl data than the other categories when deciding our multilingual data mixture.

\begin{table*}[!t]\small\centering
\renewcommand{\arraystretch}{1.2} 
\setlength{\tabcolsep}{6pt} 
\begin{tabular}{c|ccccccccc}
\toprule
\rowcolor[HTML]{FFFFFF} 
\textbf{Multilingual Data} & \textbf{Avg} & \textbf{Sp} & \textbf{Ge} & \textbf{Fr} & \textbf{Ma} & \textbf{It} & \textbf{Ja} & \textbf{Po} & \textbf{Ko} \\ 
\midrule
Common Crawl         & 37.0                          & 37.8             & 36.5            & 39.8            & 34.3              & 36.3             & 35.3              & 37.5                & 38.8            \\
FineWeb-2         & 35.1                          & 38.8             & 35.0            & 34.3            & 31.5              & 37.0             & 33.0              & 36.0                & 35.3            \\
DiverseQA-wiki & 42.1                         & 44.8             & 41.3            & 41.8            & 41.5              & 44.0             & 41.0              & 42.3                & 40.3                     \\ 
DiverseQA-crawl  & \textbf{47.0 }                        & \textbf{49.8}             & \textbf{50.8}            & \textbf{48.3}            & \textbf{46.0}              & \textbf{45.8}             & \textbf{44.5}              & \textbf{49.0}                & \textbf{42.0}           \\
\bottomrule
\end{tabular}
\caption{Comparison of multilingual datasets on the Global-MMLU Benchmark.}
\label{table:ablation_multi}
\end{table*}

\subsubsection{Fundamental Reasoning SFT-Style Data Ablation Study}
To show the effectiveness of the fundamental reasoning (FR) focused SFT-style data we introduced in Section~\ref{sec:pretrain_data}, we took the Nemotron-H-8B~\citep{nvidia2025nemotronhfamilyaccurateefficient} intermediate checkpoint trained over 14.5T tokens, and continuous pretrained it with another 100B tokens. We assigned 5\% of the 100B tokens to the newly synthesized FR-SFT data (as a replacement for Common Crawl data), and kept all other data categories the same as in the Nemotron-H-8B's phase 3 blend. We compared this model with Nemotron-H-8B, which had also been trained with 14.6T tokens. The detailed evaluation benchmarks are introduced in Section~\ref{subsec:base_model_evals}. The comparison results are shown in Table~\ref{table:ablation_sft}. The SFT-style data helped improve the Nemotron-H 8B model's performance on MMLU-Pro from 44.24 to 56.36, and also helped increase the average MATH score by around 2 points. While MMLU-Pro is a more challenging benchmark that evaluates a model's language understanding capability, it also requires the model to have excellent reasoning capability to select the correct answer out of ten choices. Our SFT data helps equip the model to select the correct answers from the other nine distractors through fundamental reasoning. We noticed no decrease in the average commonsense reasoning and average code benchmarks. 

\begin{table*}[!t]
\small
\centering
\setlength{\tabcolsep}{6pt} 
\renewcommand{\arraystretch}{1.2} 
\begin{tabular}{l|ccccc}
\toprule
\textbf{Model} & \textbf{Avg Math} & \textbf{Avg Code} & \textbf{Avg Reasoning} & \textbf{MMLU} & \textbf{MMLU-Pro} \\
\midrule
Nemotron-H 8B & 37.92 & 59.49 & 71.79 & 72.67 & 44.24 \\
\shortstack[l]{Nemotron-H 8B\\(w/ FR-SFT data)} & \textbf{39.70} & 59.61 & 71.43 & 72.98 & \textbf{56.36} \\
\bottomrule
\end{tabular}
\caption{Ablation study of the Fundamental Reasoning (FR) focused SFT-style data.}
\label{table:ablation_sft}
\end{table*}
\subsection{FP8 Recipe}\label{sec:fp8}

We used DeepSeek's FP8 training recipe for the entirety of the pretraining run~\citep{deepseekai2025deepseekv3technicalreport}. Specifically, we used E4M3 for all tensors, 128x128 quantization blocks for weights, and 1x128 tiles for the activations. Unlike Nemotron-H, we natively kept the model weights in E4M3 so that we could do the distributed optimizer's parameter all-gather operations (across data-parallel replicas) in FP8; master weights are still kept in FP32. One exception to DeepSeek's formula was that we left the first and last four linear layers in BF16, as done with Nemotron-H. Also unlike the DeepSeek-V3 run, we left all optimizer state in FP32. We observed no training instabilities from this choice of numerics.

\subsection{Hyperparameters}\label{sec:pre-train-hyperparams}

We trained \ourbasemodel on a token horizon of 20 trillion tokens. We used a sequence length of 8192 and global batch size of 768 (6,029,312 tokens per batch). We did not use any batch size ramp-up. We used a WSD (Warmup-Stable-Decay)~\citep{hu2024minicpm} learning rate schedule with a ``stable'' learning rate of $4.5\cdot10^{-4}$ and a minimum value of $4.5\cdot10^{-6}$; the learning rate was decayed over the final 3.6 trillion tokens. Weight decay was set to 0.1, and Adam $\beta_1$ and $\beta_2$ were set to 0.9 and 0.95 respectively

\subsection{Long-Context Extension}\label{sec:pre-train-long-context}

To ensure \ourbasemodel can infer over long context windows, we added a long-context phase (Phase LC) after Phase 3 of pre-training. In Phase LC, we did continuous pretraining (CPT) with a context length of 524,288 (512k) tokens using a constant learning rate of $4.5\cdot10^{-6}$. Although the target context length of \ourmodel is 128k, in preliminary studies on the Nemotron-H 8B model, we found it better to do CPT with 512k sequence length, instead of 256k or 128k. Our intuition is that longer training sequence can effectively lower the chance of long coherent documents being cut and separated by the Concat \& Chunk algorithm for pretraining data loading. We used 8-way tensor model parallelism and 16-way context parallelism to ensure training with sequence lengths of 512k tokens still fits in GPU memory. We used a global batch size of 12 to ensure the total number of tokens per global batch during long-context CPT is the same as during pretraining: around 6M tokens. Phase LC consisted of 18.9 billion tokens.

Additionally, we did long-context synthetic data generation to create more high-quality data for Phase LC. Since the academic pretraining dataset is a good source of coherent long-context documents, we used such documents that are longer than 32k tokens as seed data. We followed the methods mentioned in the Llama-3~\citep{grattafiori2024llama3herdmodels} and Qwen-2.5~\citep{qwen2025qwen25technicalreport} tech reports to generate long-context document QA data. We split each document into chunks of 1,024 tokens and then randomly selected 10\% of the chunks to be fed into Qwen-2.5-72B-Instruct for data synthesis. We asked the generator to generate a QA pair based on the information in the text chunk. We concatenated the QA pairs and appended them to the end of the original document as a sample of the long-context document QA data. Such long-document QA provided good material for the model to learn long-context dependencies. See Table~\ref{table:ablation_lc} for ablation results on Nemotron-H 8B regarding train sequence lengths and the effects of synthetic data.

The data blend used in Phase LC was built based on that of Phase 3. We proportionally downscaled the weights of all Phase 3 data to 80\% of their original values, allocating the remaining 20\% to the newly added long-context document-QA data. We found such a blend could effectively extend the context length of \ourbasemodel without degrading regular benchmark scores.

\begin{table*}[!hbt]\small\centering
\renewcommand{\arraystretch}{1.2} 
\setlength{\tabcolsep}{6pt} 
\begin{tabular}{l|cccc}
\toprule
Train length & \textbf{128k} & \textbf{256k} & \textbf{256k} & \textbf{512k}  \\
Synthetic data         & yes           & no            & yes           & yes            \\
\midrule
RULER-128k             & 73.68         & 70.19         & 79.04         & \textbf{81.04} \\
\bottomrule
\end{tabular}
\caption{Comparisons of different train sequence lengths and synthetic data usages. Ablations were conducted on Nemotron-H 8B.}
\label{table:ablation_lc}
\end{table*}

\begin{table}[!ht]
\small
\centering

\renewcommand{\arraystretch}{1.2} 

\begin{tabular}{lc|ccc}
\toprule
\multirow{2}{*}{\textbf{Task}} & \textbf{N-Nano-V2} & \textbf{N-Nano-V2} & \textbf{Qwen3 } & \textbf{Gemma3} \\
                               & \textbf{12B Base}  & \textbf{9B Base}   & \textbf{8B Base}     & \textbf{12B Base}       \\
\toprule
\textbf{General}       & \multicolumn{1}{l}{} & \multicolumn{1}{l}{} & \multicolumn{1}{l}{} & \multicolumn{1}{l}{} \\
MMLU                   & 78.24                & 74.53       & \textbf{76.44}                & 73.61                \\
MMLU-Pro 5-shot        & 63.98                & \textbf{59.43}       & 56.27                & 45.12                \\
AGIEval English CoT       & 68.03                & \textbf{65.28}       & 59.54                & 51.69                \\
\midrule
\textbf{Math}          & \multicolumn{1}{l}{} & \multicolumn{1}{l}{} & \multicolumn{1}{l}{} & \multicolumn{1}{l}{} \\
GSM8K CoT & 91.66                & \textbf{91.36}       & 84.00                & 74.45                \\
MATH           & 83.54                & \textbf{80.50}       & 55.40                & 42.40                \\
MATH Level 5   & 67.61                & \textbf{63.64}       & 29.91                & 17.71                \\
AIME 2024 pass@32      & 56.67                & \textbf{30.00}       & 20.00                & 16.67                \\
\midrule
\textbf{Code}          & \multicolumn{1}{l}{} & \multicolumn{1}{l}{} & \multicolumn{1}{l}{} & \multicolumn{1}{l}{} \\
HumanEval+ avg@32      & 61.03                & \textbf{58.50}       & 57.55                & 36.68                \\
MBPP+ avg@32           & 61.55                & \textbf{58.95}       & 58.56                & 51.73                \\
\midrule
\multicolumn{5}{l}{\textbf{Commonsense Understanding}}                                                                                    \\
ARC Challenge          & 93.26                & 90.70                & \textbf{93.09}       & 90.44                \\
HellaSwag              & 84.00                & 79.90                & 79.75                & \textbf{84.15}       \\
OpenBookQA             & 46.00                & 44.80                & 42.00                & \textbf{46.00}       \\
PIQA                   & 82.54                & 81.83                & 79.43                & \textbf{82.10}       \\
WinoGrande             & 79.24                & 75.30                & 75.93                & \textbf{79.95}       \\
\midrule
\textbf{Long Context}          & \multicolumn{1}{l}{} & \multicolumn{1}{l}{} & \multicolumn{1}{l}{} & \multicolumn{1}{l}{} \\
RULER-128K             & 84.74                & \textbf{82.22}       & -                    & 80.70                \\
\bottomrule
\end{tabular}

\caption{Accuracy of Nemotron-Nano-V2-Base models versus existing SoTA models. N-Nano-V2 is short for Nemotron-Nano-V2. The distilled N-Nano-V2-9B-Base is compared against Qwen3-8B-Base and Gemma3-12B-Base, and the best score is highlighted in each row.}
\label{table:base_evals}
\end{table}

\subsection{Base Model Evaluations}
\label{subsec:base_model_evals}

We run evaluations of all models ourselves unless otherwise stated. Our evaluation setup is built on top of \texttt{lm-evaluation-harness}\footnote{\url{https://github.com/EleutherAI/lm-evaluation-harness}.} for fair comparisons, with the following changes:
\begin{enumerate}
    \item For mathematical reasoning, we evaluate GSM8K and MATH~\citep{cobbe2021trainingverifierssolvemath, hendrycks2021measuringmathematicalproblemsolving} benchmarks using greedy-decoding. We also highlight the competition-level slice of the MATH benchmark as ``MATH Level 5''. Additionally, we report the $\text{pass}@32$ performance on AIME-2024. We use \texttt{Math-Verify}\footnote{\url{https://github.com/huggingface/math-verify}.} to grade all generations.
    \item For code tasks (HumanEval~\citep{chen2021evaluatinglargelanguagemodels}, MBPP~\citep{austin2021programsynthesislargelanguage}) we evaluate the EvalPlus variants along with the sanitization of generations~\citep{Liu_Is_Your_Code_2023}, in a 0-shot setup. We estimate $\text{avg}@32$,  $\text{pass}@1$ from 32 generations per prompt.
    \item General reasoning benchmarks (OpenBookQA~\citep{mihaylov2018suitarmorconductelectricity}, PIQA~\citep{bisk2019piqareasoningphysicalcommonsense}, Hellaswag~\citep{zellers2019hellaswagmachinereallyfinish}, Winogrande~\cite{sakaguchi2019winograndeadversarialwinogradschema}) are unchanged except for ARC-Challenge~\citep{Clark2018ThinkYH}, where we present all options at the same time, similar to MMLU~\citep{hendrycks2021measuringmassivemultitasklanguage}.
    \item For multilingual capability, we evaluate MGSM~\cite{shi2022languagemodelsmultilingualchainofthought} (8-shot, native CoT) and Global MMLU-Lite~\cite{singh2024globalmmluunderstandingaddressing}.
    \item We use RULER \citep{hsieh2024ruler} as the long context benchmark. We report the average scores over all the 13 tasks included in RULER.
\end{enumerate}

Accuracy results for \ourbasemodel with comparsions to Qwen3-8B Base and Gemma3-12B Base are shown in Tables~\ref{table:base_evals} and~\ref{table:base_evals_multilingual}. We also include the accuracy of our 9B pruned variant of \ourbasemodel which is discussed in Section~\ref{sec:compression}.

\begin{table}[!ht] \small\centering
\renewcommand{\arraystretch}{1.2} 
\begin{tabular}{lc|ccc}
\toprule
\multirow{2}{*}{\textbf{Task}} & \textbf{N-Nano-V2} & \textbf{N-Nano-V2} & \textbf{Qwen3 } & \textbf{Gemma3} \\
                               & \textbf{12B Base}  & \textbf{9B Base}   & \textbf{8B Base}     & \textbf{12B Base}       \\
\toprule
\multicolumn{5}{l}{\textbf{Global-MMLU-Lite}}              \\
German      & 74.50 & 68.25  & \textbf{75.50}  & 69.75 \\
Spanish     & 76.50 & 72.75  & \textbf{75.00}  & 74.00 \\
French      & 78.25 & 69.75  & \textbf{74.25}  & 72.50 \\
Italian     & 76.50 & 73.25  & 72.75  & \textbf{74.00} \\
Japanese    & 71.00 & 67.00  & 70.00  & \textbf{71.50} \\
Korean      & 72.50 & 67.25  & 67.25  & \textbf{70.25} \\
Portuguese  & 76.25 & 71.25  & 72.50  & \textbf{75.75} \\
Chinese     & 75.50 & 69.25  & \textbf{75.25}  & 67.25 \\
Average     & 75.13 & 69.94  & \textbf{72.81}  & 71.88 \\
\midrule
\multicolumn{5}{l}{\textbf{Multilingual Math (MGSM)}} \\
Spanish     & 93.20 & \textbf{93.60}  & 87.60  & 73.60 \\
German      & 88.40 & \textbf{88.40}  & 78.80  & 66.00 \\
French      & 82.40 & \textbf{84.40}  & 82.00  & 68.00 \\
Chinese     & 83.60 & \textbf{82.00}  & 80.80  & 62.00 \\
Japanese    & 76.80 & 68.80  & \textbf{71.20}  & 56.00 \\
Russian     & 91.20 & \textbf{90.80}  & 85.20  & 72.40 \\
Average     & 85.94 & \textbf{84.67}  & 80.93  & 66.33 \\
\bottomrule
\end{tabular}
\caption{Accuracy of Nemotron-Nano-V2-Base models versus existing SoTA models on multilingual benchmarks. N-Nano-V2 is short for Nemotron-Nano-V2. The distilled N-Nano-V2-9B-Base is compared against Qwen3-8B-Base and Gemma3-12B-Base, and the best score is highlighted in each row.}
\label{table:base_evals_multilingual}
\end{table}

\section{Alignment}
\label{sec:alignment}
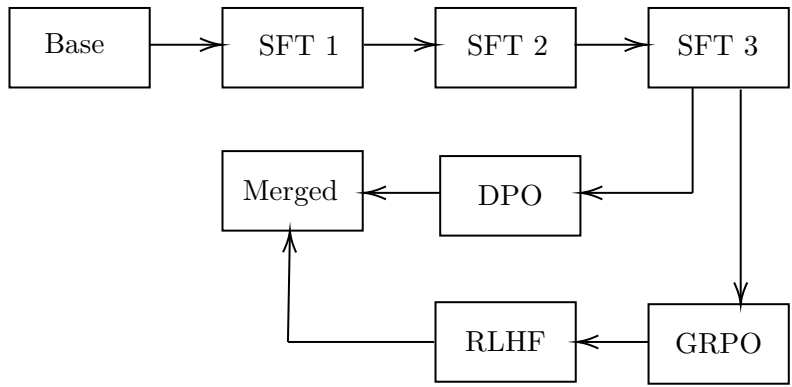
\begin{figure}[h]
    \centering

\tikzset{
  every picture/.style={line width=0.75pt}, 
  boxlabel/.style={
    align=center, inner sep=0pt, outer sep=0pt,
    text height=1.7ex, text depth=.3ex, 
    font=\normalsize                       
  }
}

\begin{tikzpicture}[x=0.75pt,y=0.75pt,yscale=-1,xscale=1]

\draw (206.33,40) -- (276.33,40) -- (276.33,80) -- (206.33,80) -- cycle;   
\draw (312.66,40) -- (382.66,40) -- (382.66,80) -- (312.66,80) -- cycle;   
\draw (100,40)    -- (170,40)    -- (170,80)    -- (100,80)    -- cycle;   
\draw (419,40)    -- (489,40)    -- (489,80)    -- (419,80)    -- cycle;   

\draw (315,114.5) -- (385,114.5) -- (385,154.5) -- (315,154.5) -- cycle;   
\draw (206.33,112) -- (276.33,112) -- (276.33,152) -- (206.33,152) -- cycle; 

\draw (419,189) -- (489,189) -- (489,229) -- (419,229) -- cycle;           
\draw (312.66,188) -- (382.66,188) -- (382.66,228) -- (312.66,228) -- cycle; 

\draw (170,58.5) -- (204,58.5);
\draw[shift={(206,58.5)}, rotate=180] (10.93,-3.29) .. controls (6.95,-1.4) and (3.31,-0.3) .. (0,0) .. controls (3.31,0.3) and (6.95,1.4) .. (10.93,3.29);

\draw (277,58.5) -- (311,58.5);
\draw[shift={(313,58.5)}, rotate=180] (10.93,-3.29) .. controls (6.95,-1.4) and (3.31,-0.3) .. (0,0) .. controls (3.31,0.3) and (6.95,1.4) .. (10.93,3.29);

\draw (382,58.5) -- (416,58.5);
\draw[shift={(418,58.5)}, rotate=180] (10.93,-3.29) .. controls (6.95,-1.4) and (3.31,-0.3) .. (0,0) .. controls (3.31,0.3) and (6.95,1.4) .. (10.93,3.29);

\draw (465,81) -- (465,187);
\draw[shift={(465,189)}, rotate=270] (10.93,-3.29) .. controls (6.95,-1.4) and (3.31,-0.3) .. (0,0) .. controls (3.31,0.3) and (6.95,1.4) .. (10.93,3.29);

\draw (419,208) -- (385,208);
\draw[shift={(383,208)}, rotate=360] (10.93,-3.29) .. controls (6.95,-1.4) and (3.31,-0.3) .. (0,0) .. controls (3.31,0.3) and (6.95,1.4) .. (10.93,3.29);

\draw (312,208) -- (239,208);
\draw (239,208) -- (239.96,154);
\draw[shift={(240,152)}, rotate=91.02] (10.93,-3.29) .. controls (6.95,-1.4) and (3.31,-0.3) .. (0,0) .. controls (3.31,0.3) and (6.95,1.4) .. (10.93,3.29);

\draw (441,80) -- (441,133);
\draw (441,133) -- (387,133);
\draw[shift={(385,133)}, rotate=360] (10.93,-3.29) .. controls (6.95,-1.4) and (3.31,-0.3) .. (0,0) .. controls (3.31,0.3) and (6.95,1.4) .. (10.93,3.29);

\draw (315,133) -- (279,133);
\draw[shift={(277,133)}, rotate=360] (10.93,-3.29) .. controls (6.95,-1.4) and (3.31,-0.3) .. (0,0) .. controls (3.31,0.3) and (6.95,1.4) .. (10.93,3.29);

\draw (116,51) node[anchor=north west, inner sep=0.75pt, align=left] {Base};
\draw (223,52) node[anchor=north west, inner sep=0.75pt, align=left] {SFT 1};
\draw (327,52) node[anchor=north west, inner sep=0.75pt, align=left] {SFT 2};
\draw (432,52) node[anchor=north west, inner sep=0.75pt, align=left] {SFT 3};

\draw (215,125) node[anchor=north west, inner sep=0.75pt, align=left] {Merged};

\node[boxlabel] at ($(419,189)!0.5!(489,229)$)        {\strut GRPO};
\node[boxlabel] at ($(312.66,188)!0.5!(382.66,228)$)  {\strut RLHF};
\node[boxlabel] at ($(315,114.5)!0.5!(385,154.5)$)    {\strut DPO};

\end{tikzpicture}
    \caption{Flow of alignment procedures followed to arrive at the final "Merged" Nemotron Nano 2 12B checkpoint.}
    \label{fig:alignflow}
\end{figure}

In this section we will present the alignment process we followed to convert the base checkpoint into an aligned 12B checkpoint. Our process is outlined in Figure \ref{fig:alignflow}.
\subsection{Post-Training Data}\label{sec:sftdata}
Our alignment begins with a large-scale SFT stage which trains the base model on approximately 80 billion tokens of prompt-response pairs. The distribution of domains is shown in Table~\ref{tab:domain_data}. 
\begin{table}[h]
\centering
\begin{tabular}{l r}
\toprule
\textbf{Domain} & \textbf{$~$Number of Samples} \\
\toprule
Math                  & 1.5M \\
Coding                & 1.1M \\
Science               & 2.0M \\
Tool-calling          & 400K \\
Conversational        & 1.5M \\
Safety                & 2K \\
Multilingual (all domains) & 5.0M \\
\bottomrule
\end{tabular}
\caption{Post-training data distribution across domains used for our SFT stages.}
\label{tab:domain_data}
\end{table}

\paragraph{Math, science and coding.} For Math~\citep{toshniwal2024openmath2, moshkov2025aimo2}, Science and Coding~\citep{ahmad2025opencodereasoning,ahmad2025opencodeinstruct,majumdar2024genetic} data, we generate responses using the open-weights DeepSeek-R1-0528 model~\citep{deepseekai2025deepseekv3technicalreport} using the same prompts used for training Nemotron-H-8B and 47B Reasoning models~\citep{nvidia2025nemotronhfamilyaccurateefficient}. The training data has been released as part of {\texttt Nemotron-Post-Training-Dataset-v1}\footnote{\url{https://huggingface.co/datasets/nvidia/Nemotron-Post-Training-Dataset-v1}}.

\paragraph{Tool calling.} The tool-calling dataset consists of single-turn, multi-turn, and multi-step conversations. 
For single-turn cases, we sample prompts from \texttt{xlam-function-calling-60k}\footnote{\url{https://huggingface.co/datasets/xlam-function-calling-60k}}, 
\texttt{glaive-} \texttt{function-calling-v2}\footnote{\url{https://huggingface.co/datasets/glaive-function-calling-v2}}, 
\texttt{NVIDIA-When2Call}~\citep{Ross2025When2Call}, 
and generate responses using \texttt{Qwen3-235B-A22B}\footnote{\url{https://huggingface.co/Qwen/Qwen3-235B-A22B}}. 
Inspired by ToolACE~\citep{Liu2024ToolACE} and APIGen-MT~\citep{Prabhakar2025APIGenMT}, 
we extend this to multi-turn and multi-step settings by simulating conversations where \texttt{Qwen3-235B-A22B} plays the roles of User-Agent, Assistant-Agent, and API-Server-Agent. 
The User-Agent reviews available tools, poses challenging queries, interacts when addressed by the Assistant, and judges task success at the end. 
Each instance is paired with a random persona from \texttt{Nemotron-Personas}\footnote{\url{https://huggingface.co/datasets/NVIDIA/Nemotron-Personas}} to enrich diversity of queries. 

The Assistant-Agent receives the initial query and available tools, executes tasks by invoking tools, interpreting their responses, and interacting with the User-Agent across single-turn, multi-turn, or multi-step scenarios. 
Meanwhile, the API-Server-Agent acts as a mock API server, checking parameters and returning either valid outputs or error messages depending on correctness. 
A lightweight rule-based tool-call verification layer further strengthens reliability by ensuring outputs are consistent and verifiable, and only successful trajectories are retained. 

\paragraph{Multilingual data.} Our multilingual synthetic post-training data are constructed by translating existing English post-training data. To address the challenges of Large Language Model (LLM) hallucinations and quality degradation on long inputs when generating synthetic translation data, we implement a robust quality assurance pipeline. Our method involves translating inputs line-by-line to manage complexity and skip non-translatable content like code. We also enforce a strict bracket format for reliable extraction and use language identification to filter out off-target translations, thereby ensuring high-quality final outputs.

\paragraph{Conversational data.} For conversational data, we use prompts from the LMSYS dataset~\citep{zheng2023judging} and generate responses using the \texttt{Qwen3-235B-A22B} reasoning model~\citep{yang2025qwen3technicalreport}. 
We also incorporate prompts from HelpSteer2 and HelpSteer3, paired with responses generated by the same model. 
In addition, we draw on a subset of approximately 550k prompts from WildChat-1M~\citep{li2024wildchat}, again generating reasoning responses with \texttt{Qwen3-235B-A22B}. We also include multi-turn conversations with Deepseek R1 responses using the multi-turn conversational prompts used in~\cite{nvidia2025nemotronhfamilyaccurateefficient}.

\paragraph{Safety.} We leveraged a mix of harmful and benign prompts drawn from the Nemotron Content Safety Dataset V2~\citep{ghosh-etal-2025-aegis2}\footnote{\url{https://huggingface.co/datasets/nvidia/Aegis-AI-Content-Safety-Dataset-2.0}}, 
HarmfulTasks~\citep{hasan2024pruning}, 
RedTeam2K~\citep{luo2024jailbreakv}, 
and gretel-v1~\citep{gretelai_gretel-safety-alignment-en-v1}. 
Responses were generated using DeepSeek-R1-0528\footnote{\url{https://huggingface.co/deepseek-ai/DeepSeek-R1}}. 
To ensure safety, we applied a two-step approach: initial prompting followed by filtering with guard models to verify that outputs remained safe.

\subsection{Post Training}
\label{sec:supervised_fine_tuning}
\paragraph{Stage 1 SFT.} As Figure~\ref{fig:alignflow} illustrates, we employ three distinct stages of supervised fine-tuning. 
Stage~1 uses the full dataset described in Section~\ref{sec:sftdata}, augmented with a subsample of roughly 10\% of prompts paired with outputs stripped of reasoning traces. 
This exposes the model to ``empty'' traces, enabling it to produce direct answers in a reasoning-off mode. 
To improve efficiency and preserve long-context ability from pretraining, we concatenate samples into sequences of approximately $128$k tokens, reducing padding overhead and encouraging long-range learning.

\paragraph{Stage 2 SFT.} Stage~2 targets tool-calling. 
Although Stage~1 improved performance on most benchmarks, tool-calling accuracy degraded. 
We attribute this to sample concatenation at $128$k, which likely disrupted learning of tool-calling patterns. 
Thus, Stage~2 was trained without concatenation, using the full tool-calling dataset and a representative subsample of other domains.

\paragraph{Stage 3 SFT.} Stage~3 reinforces long-context capability. 
It incorporates long-context data following the recipe used in Nemotron-H preparation~\citep{nvidia2025nemotronhfamilyaccurateefficient}, along with augmented examples across domains where reasoning traces were abruptly truncated to 1--2k tokens while preserving the final answer. 
This truncation strategy improved robustness under varying inference-time thinking budgets.
\paragraph{IFeval RL.} To improve instruction adherence, we sampled 16,000 prompts from the LMSYS Chat dataset and augmented them with IFEval-style instructions. A rule-based verifier scored outputs based on how well they satisfied each instruction, creating a reward signal that prioritized following directions with precision. IFEval RL experiments provided significant boost to IFEval capabilities while the rest of the benchmarks fluctuated slightly requiring careful checkpoint selection. 

\paragraph{DPO.}In another branch of training, we apply the DPO algorithm to improve tool-calling. 
We evaluate performance using the BFCL v3 benchmark, which extends BFCL v2 with greater emphasis on multi-step (multiple tool calls to achieve a goal) and multi-turn (multiple user–agent interactions). 
To strengthen these capabilities in the Nano V2 aligned model, we use the WorkBench environment, a multi-step verifiable tool-calling setup adapted from Styles~\citep{Styles2024WorkBench}. 
In each WorkBench task, the model must issue a sequence of tool calls across multiple steps, with correctness verified through database state comparisons. 

Nano V2 undergoes reinforcement learning in this environment through iterative stages of Direct Preference Optimization. 
For each candidate checkpoint from the long-context stage, we generate on-policy data consisting of positive examples (successful tool calls) and negative examples (failed generations) for every WorkBench prompt. 
This process ensures that iterative DPO remains on-policy.

\paragraph{RLHF.}
We evaluate the model's overall helpfulness and chat capabilities using the Arena-Hard benchmark. To improve performance on this benchmark, we use GRPO to train candidate checkpoints from the SFT stage using English-only contexts from  HelpSteer3~\citep{Wang2025HelpSteer3Preference}. During training, we generate responses both with and without thinking traces and use a Qwen-based reward model to judge the rollouts.
\paragraph{Model Merging.}
During training, we observed a trade-off between reasoning capabilities and chat capabilities. 
To address this, we opted for checkpoint interpolation~\cite{wortsman2022modelsoup}, blending in an RL checkpoint with strong reasoning capabilities with an RL checkpoint with strong chat capabilities. Checkpoint interpolation is performed by linearly interpolating model weights: $(1-\alpha) \cdot w_{model1} + \alpha \cdot w_{model2}$. We experimented with a parameter sweep over $\alpha$ values from 0.1 to 0.9 in increments of 0.1, and found that values around 0.5 offered a good trade-off.

\subsection{Evaluation}

Our 12B model's performance is summarized in Table~\ref{tab:evaluation_results}.
To test reasoning capabilities across domains, we evaluate the models on \textsc{MATH-500} \citep{lightman2023lets}, \textsc{AIME-2024}, \textsc{AIME-2025}, \textsc{GPQA-Diamond} \citep{rein2023gpqa}, \textsc{LiveCodeBench (07/24 - 12/24)} \citep{jain2024livecodebench}, \textsc{SciCode} \citep{tian2024scicoderesearchcodingbenchmark}, and \textsc{Humanity's Last Exam} \citep{phan2025humanitysexam}. For broader evaluation on diverse capabilities, we use \textsc{IFEval} \citep{zhou2023instruction} for instruction following capabilities, \textsc{BFCL v3} \citep{berkeley-function-calling-leaderboard} for tool-calling, \textsc{RULER} for long-context, and \textsc{ArenaHard} \citep{arenahard2024} for chat capability.

We conduct evaluations using NeMo-Skills\footnote{\tiny \url{https://github.com/NVIDIA/NeMo-Skills}}. We report \textsc{Pass@1} average of 16 runs for \textsc{AIME-2024}, \textsc{AIME-2025}; average of 4 runs for \textsc{MATH-500}, \textsc{GPQA-Diamond}, \textsc{LiveCodeBench}, \textsc{IFEval}; and score of 1 run for \textsc{BFCL v3}, \textsc{SciCode}, \textsc{Humanity's Last Exam}, \textsc{RULER}, and \textsc{ArenaHard}.
\begin{table*}[t]
\centering
\footnotesize
\setlength{\tabcolsep}{8pt}
\renewcommand{\arraystretch}{1.2}
\begin{tabular}{lccc}
\toprule
\textbf{Evaluation} & \textbf{Nemotron-Nano-v2-12B} & \textbf{Qwen3-8B} & \textbf{Qwen3-14B} \\
\toprule
\textsc{AIME-2024} & 85.42 & 75.83 & 81.53 \\
\textsc{AIME-2025} & 76.25 & 69.31 & 66.6 \\
\textsc{MATH-500} & 97.75 & 96.3 & 96.85 \\
\textsc{GPQA-Diamond}  & 64.48 & 59.61 & 64.53 \\
\textsc{LiveCodeBench (07/24--12/24)} & 70.79 & 59.5 & 63.08 \\
\textsc{SciCode Sub-Task} & 18.75 & 24.65 & 26.04 \\
\textsc{Humanity's Last Exam} & 6.30 & 4.40 & 5.38 \\
\textsc{IFEval (Inst. Strict)} & 89.81 & 89.39 & 91.32 \\
\textsc{BFCL v3} & 66.98 & 66.34 & 68.01 \\
\textsc{RULER @ 128k} & 83.36 & 74.13 & 73.55 \\
\textsc{ArenaHard} & 74 & 78.4 & 87.7 \\
\bottomrule
\end{tabular}
\caption{Evaluation results with reasoning "ON" (for \textbf{Nemotron-Nano-v2-12B}, \textbf{Qwen3-8B}, and \textbf{Qwen3-14B} across reasoning and general capability benchmarks.}
\label{tab:evaluation_results}
\end{table*}
 \subsection{Budget Control Evaluation}\label{sec:budgetcontrol}
Nemotron Nano V2 allows users to specify how many thinking tokens the model may generate before producing the final answer. 
The final answer is the portion of text typically shown to end users. 
This feature is implemented by counting tokens after the model begins generating the \verb|<think>| token. 
Once the budget is reached, the inference setup attempts to insert a closing \verb|</think>| tag. 
Rather than inserting it immediately, we let the model finish its current sentence and place the tag at the next newline. 
In extreme cases where no newline appears, the system enforces closure within 500 tokens past the budget: if no newline occurs by the $(\text{budget} + 500)$\textsuperscript{th} token, the \verb|</think>| tag is forcibly inserted.
\begin{figure}[ht]
    \centering
    \begin{subfigure}{0.8\linewidth}
        \centering
        \includegraphics[width=\linewidth]{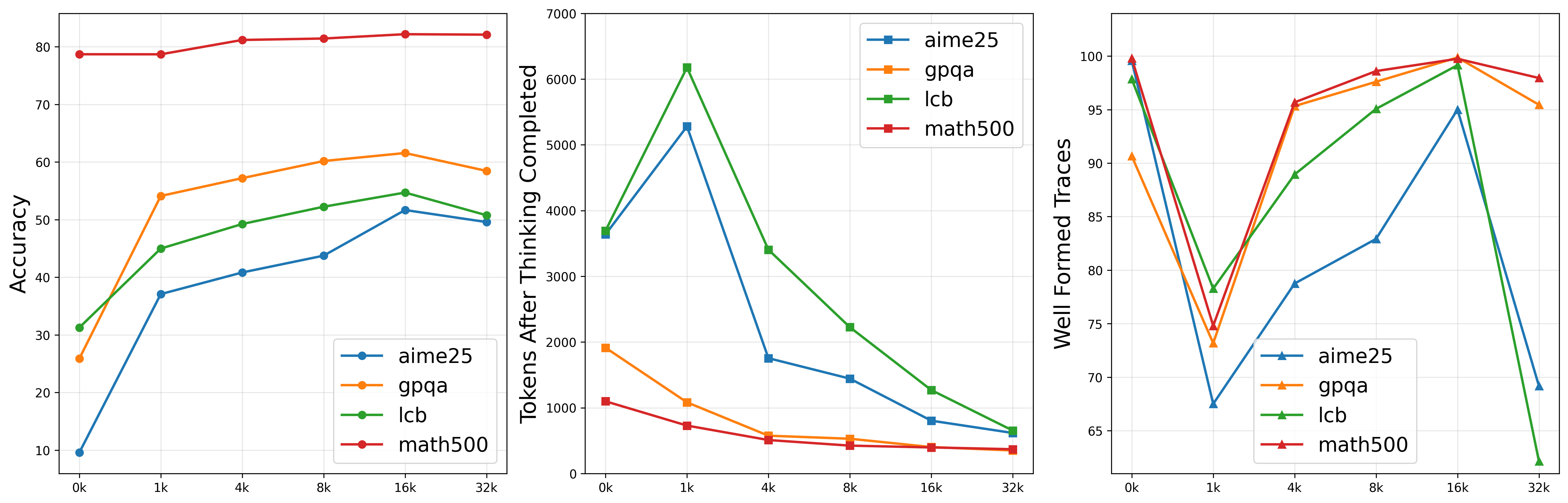}
        \caption{}
        \label{fig:badbudget}
    \end{subfigure}
    
    \vspace{0.5em} 
    
    \begin{subfigure}{0.8\linewidth}
        \centering
        \includegraphics[width=\linewidth]{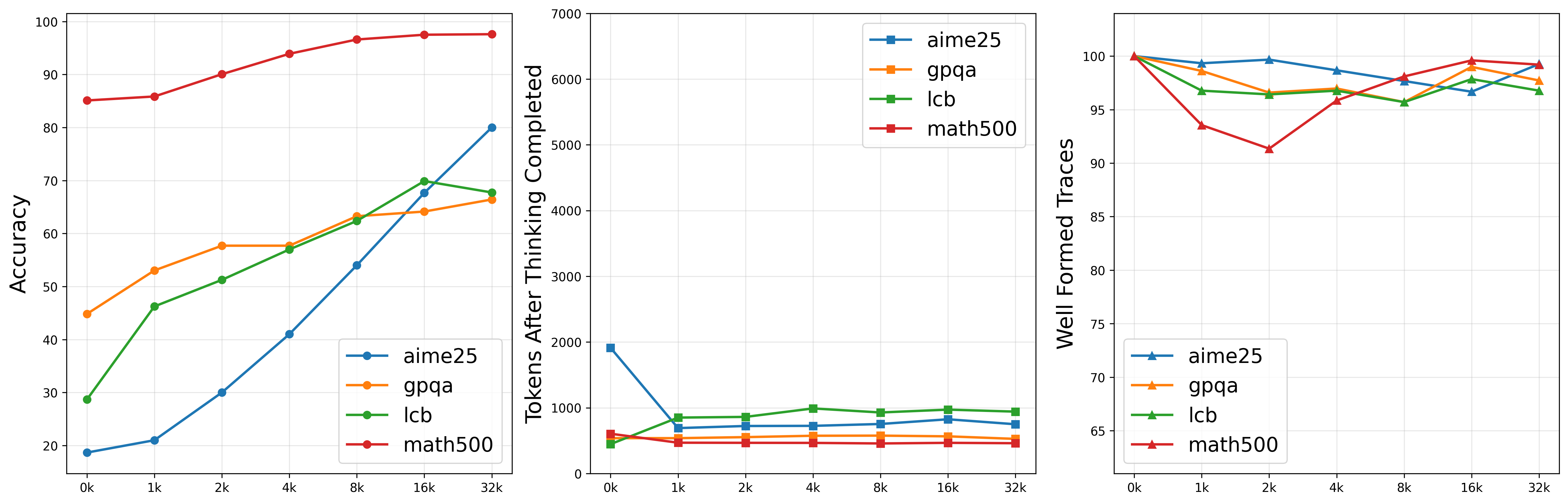}
        \caption{}
        \label{fig:budget}
    \end{subfigure}
    
    \caption{Comparison of budget control before truncation training (a) and after truncation training was included (b). For all plots above the x-axis indicates the budget assigned for thinking tokens.}
    \label{fig:budget_comparison}
\end{figure}
Figure~\ref{fig:budget} shows our models budget control behavior. Apart from just presenting the accuracy of the model at various budgets, we also inspect if the model generations are well-formatted at various budgets. We inspect for two kinds of failure modes:
\begin{itemize}

    \item In one failure mode, the model uses more tokens in the final answer to ``compensate'' for restrictions in the thinking traces. Without truncated training examples in the SFT stage, this compensation effect is prevalent (Figure~\ref{fig:badbudget}, center). With truncated training, however, the effect is absent (Figure~\ref{fig:budget}, center).
    
    \item Another issue is that the model can remain in ``thinking mode'' even after the closing tag \verb|</think>| is inserted. This is evident when the model generates the closing tag again after the forced insertion, suggesting it does not fully ``register'' the artificial closure. We evaluate this using ``Well-Formedness,'' where a well-formed response should contain only a single closing tag (either forced by the budget or produced naturally). Figure~\ref{fig:badbudget} (right) shows that for short budgets, the percentage of well-formed responses drops sharply. With truncation training, however, the model consistently produces well-formed responses (Figure~\ref{fig:budget}, right).
\end{itemize}
\section{Pruning and Distillation}
\label{sec:compression}

In this section, we describe the pruning and distillation process to compress the aligned 12B model to the Nano 2 model with the goal of running longer context (128k sequence length) inference on the NVIDIA A10G GPU.
Note that storing just the weights of a 12B parameter model in \texttt{bfloat16} precision requires 22.9 GiB, which is more than the 22 GiB memory capacity of an A10G GPU; this clearly indicates the need for compression. 

Our compression strategy builds on Minitron ~\citep{muralidharan2024compactlanguagemodelspruning,sreenivas2024llmpruningdistillationpractice,taghibakhshi2025efficient}, which is a lightweight model pruning framework for LLMs. While Minitron was originally designed for compressing pretrained base models targeting user-defined parameter budgets, in this work, we extend it to compress reasoning models while also incorporating the memory constraints and throughput-based objectives stated above.

\subsection{Importance Estimation}
\label{sec:importance}

We collect importance or sensitivity scores for each model component (e.g., layers, FFN neurons) to help decide which components to remove; this is the {\em importance estimation} phase.
The scores computed in this phase are used to decide which model components can be pruned.
We note that sensitivity analysis based on gradient information is typically impractical at modern LLM scale~\citep{muralidharan2024compactlanguagemodelspruning};
instead, we rely on a lightweight strategy that uses only forward passes. In this work, we use a simplified approach that works well in our ablation studies: a) prune layers, and b) prune FFN hidden dimensions (effectively neurons) and embedding channels.
We also experimented with pruning Mamba heads; unfortunately, this axis caused severe accuracy degradation.
We now describe how we compute the importance of each layer, embedding channel, FFN neuron and Mamba head.

\paragraph{Layer importance.}
We compute layer importance in an iterative fashion: for each candidate layer, we temporarily remove it from the model and compute the mean squared error (MSE) between the original model’s logits and those produced by the pruned model. This MSE reflects the contribution of that layer to the model’s predictions: lower values indicate smaller impact. At each pruning step, we remove the layer with the lowest MSE, as it has the least influence on the final output. We repeat this process until the desired depth is reached. This strategy ensures that pruning preferentially removes layers whose absence minimally affects the model’s behavior. For more details on iterative MSE-based layer importance, please refer to~\citet{nvidia2025nemotronhfamilyaccurateefficient}.

\paragraph{FFN and embedding channel importance.}\label{mlp_pruning} FFN layers internally are composed of two linear operators with a non-linear activation in between: 
$$
\operatorname{FFN}(\mathbf{X}) = \delta\bigg(\mathbf{X} \cdot \boldsymbol{W}^{T}_{1}\bigg) \cdot \boldsymbol{W}_{2}.
$$
Here, $\mathbf{X}$ denotes the input, and $\boldsymbol{W}_{1}$ and $\boldsymbol{W}_{2}$ are the two associated weight matrices in the FFN layer. 
$\boldsymbol{W}_{1}, \boldsymbol{W}_{2} \in \mathbb{R}^{d_{ffn}\times d_{model}}$, where $d_{model}$ and $d_{ffn}$ are the model hidden dimension and FFN hidden dimension respectively.
$\delta(\cdot)$ refers to the non-linear activation function (squared ReLU in this work).

Following the same procedure as Minitron~\citep{muralidharan2024compactlanguagemodelspruning}, we compute the importance of each neuron in the first linear operator of each FFN layer by examining the set of outputs it produces. We use a small calibration dataset of 1024 samples for this purpose. Formally, we compute each neuron's importance score by aggregating its outputs given an input batch $X$:
$$F_{\text{neuron}}^{(i)} = \sum_{\mathbf{B,S}} \delta\bigg(\mathbf{X} \big(\boldsymbol{W}_{1}^{i}\big)^T\bigg).$$
Here, $\boldsymbol{W}_{1}^{i}$ refers to the $i^\text{th}$ row of the weight matrix $\boldsymbol{W_{1}}$. 
$\sum_{\mathbf{B, S}}$ refers to aggregation along the batch and sequence dimensions. 
We use the \texttt{mean} and \texttt{l2-norm} aggregation functions along the batch and sequence dimensions, following the observations in the Minitron paper.
For a sequence of scores $\mathbf{S}$, \texttt{mean} aggregation is defined as $\frac{1}{n}\sum_{i=1}^{n}|\mathbf{S}_i|$, and \texttt{l2-norm} is $\sqrt{\sum_{i=1}^{n}\mathbf{S}_i^2}$. 
Embedding channel importance is computed similarly, by examining the outputs of LayerNorm layers instead; we refer the reader to~\citet{muralidharan2024compactlanguagemodelspruning} for more details.

\paragraph{Mamba importance.}  
Mamba layers process inputs through multiple projection matrices 
(\(W_x,\allowbreak W_z,\allowbreak W_B,\allowbreak W_C,\allowbreak W_{dt}\)) 
that produce intermediate representations before causal convolution and selective state space model (SSM) updates, followed by gated normalization and an output projection (\(W_O\)). We follow the methodology described in~\citet{taghibakhshi2025efficient} for importance estimation: specifically, we adopt a nested activation-based scoring strategy over a small calibration dataset of 1024 samples, similar to FFN importance but adapted to Mamba’s group-aware structure. First, we obtain activation scores from the \(W_x\) projection, denoted \(s \in \mathbb{R}^{m_h \times m_d}\), where \(m_h\) is the number of Mamba heads and \(m_d\) is the Mamba head channel dimension. For each channel \(d\), the score is computed as
\begin{equation*}
s_d = \left\lVert \sum_{\mathbf{B},\mathbf{S}} s_{:,d} \right\rVert_2,
\end{equation*}
where the aggregation is over the batch (\(\mathbf{B}\)) and sequence (\(\mathbf{S}\)) dimensions, using both \texttt{mean} and \texttt{l2-norm} metrics. Next, head scores are computed by using the \texttt{l2-norm} over the Mamba head channel set:
\begin{equation*}
f_h = \left\lVert s_{h,m_d} \right\rVert_2 , \quad \forall h \in \{1,\dots,m_h\},
\end{equation*}
and heads are ranked within each Mamba group \(\mathcal{G}_g\) to preserve group-aware computation semantics:
\begin{equation*}
\mathcal{R}_g = \mathrm{argsort}_{h \in \mathcal{G}_g}(f_h).
\end{equation*}
which ensures that pruning decisions respect the model’s structural constraints and SSM's sequence modeling. The lowest-scoring heads are pruned by trimming the corresponding rows from all affected projection, convolution, and SSM parameter matrices. This strategy preserves the integrity of the SSM block while removing less important Mamba heads. As shown in~\cite{taghibakhshi2025efficient}, pruning Mamba heads yields a better accuracy–throughput trade-off than pruning head channels; we consequently focus on head pruning in this work.

\subsection{Lightweight Neural Architecture Search}
\label{sec:nas}
We first define the constraints and objectives for the Nano 2 model, and then describe our lightweight Neural Architecture Search (NAS) framework that finds the most promising architectural candidates that meet our objectives and constraints.

\paragraph{Memory constraints.}
Memory requirements during inference consist of two distinct components with different scaling behaviors. The parameter memory, while substantial, remains constant regardless of the input size. In contrast, the key-value cache
memory scales linearly with both batch size and sequence length, often becoming the dominant factor in long-sequence scenarios. For the Nano 2 model, our goal was to be able to perform inference at a sequence length of 128k and a batch size of at least 1 within a memory budget of 19.66 GiB. We obtained the budget as follows: from the 22.06 GiB available memory on an NVIDIA A10G GPU, we subtract a 5\% buffer for frameworks such as vLLM and TensorRT-LLM and another 1.3 GiB to allow sufficient space for a vision encoder.

\paragraph{Measuring throughput.}
For the experiments below, unless otherwise specified, we measure throughput on an input and output sequence length of 8k and 16k tokens respectively, which we believe represents a typical reasoning scenario. For this combination of input and output sequence length, we report vLLM output token generation throughput at the maximum batch size that fits on the A10G GPU. 

\subsubsection{Candidate enumeration.}
Our compression strategy explores multiple axes within the 19.66 GiB memory budget through combinatorial pruning. Our search space includes depth reduction (removing 6-10 layers from the original 62-layer architecture) combined with width pruning of embedding channels (4480-5120), FFN dimension (13440-20480), and Mamba heads (112-128). This multi-axis search space results in hundreds of candidate architectures meeting the memory constraint.

\subsubsection{Finding the Best Architecture}
Since performing knowledge distillation and throughput benchmarking on the full set of candidates would be prohibitively expensive, we break down the problem into two parts: (1) find the optimal depth for the compressed model, and (2) find the optimal width-pruned architecture given the depth.

\paragraph{Effect of depth.}
We compare the accuracy of three depth-pruned candidates obtained from the 12B model with 52, 54 and 56 layers. Here, we keep the number of attention layers fixed at 4 for all three variants so as to achieve a good balance between KV cache size  and long-context performance; prior work has indicated that an attention-to-total-layers ratio between 7-8\% is reasonable~\citep{nvidia2025nemotronhfamilyaccurateefficient}. We leave the width dimensions untouched for this experiment.
Table~\ref{tab:depth_ablation} lists average reasoning accuracy at different depths after 6B tokens of distillation; in line with our previous observations on the strong correlation between depth and task performance~\citep{muralidharan2024compactlanguagemodelspruning,sreenivas2024llmpruningdistillationpractice}, we notice that reducing depth beyond 56 layers results in significant accuracy degradation; as a result, we fix the depth at 56 for further width pruning.

\begin{table*}[!hbt]\small\centering
\setlength{\tabcolsep}{5pt}
\begin{tabular}{c|c}
\toprule
\textbf{} & \textbf{Accuracy (Avg)} \\ 
\midrule
52 Layers & 44.92 \\
54 Layers & 47.35 \\
56 Layers & 51.48 \\ 
\bottomrule
\end{tabular}
\caption{Effect of depth on reasoning accuracy. Results are after distilling with 6B tokens.}
\label{tab:depth_ablation}
\end{table*}

\paragraph{Combining depth and width pruning.}
As described above, we fix the depth of our target model to 56 layers with 4 attention layers. We perform 60B tokens of distillation on this checkpoint (see Section~\ref{sec:distillation} for additional details) and perform further width pruning along the embedding, FFN, and Mamba axes. We enumerate all candidate pruned architectures that meet our memory budget, and sort them in decreasing order of estimated memory consumption at 128k context length and batch size 1. The top 3 candidates from this list are picked for further evaluation: in particular, we perform short Knowledge Distillation (KD) on these candidates for 19B tokens after depth+width pruning; we also benchmark throughput to pick the final architectural candidate. Table~\ref{tab:top3} lists the architectural details of the top 3 candidates, along with the achieved task performance (post KD) and throughput. As shown in the Table, Candidate 2 achieves the best accuracy while still having reasonable runtime performance; consequently, we use this architecture for Nano 2.

\begin{table*}[!hbt]
\small\centering
\setlength{\tabcolsep}{3pt} 
\begin{tabular}{c|ccccccc}
\toprule
\multicolumn{1}{l|}{} & \textbf{\#Layers} & \textbf{Hidden} & \textbf{FFN} & \textbf{Mamba \#Heads} & \textbf{Params.~(B)} & \textbf{Accuracy} & \textbf{Throughput} \\ 
\midrule
Candidate 1 & 56 & 4480 & 17920  & 112 & 8.92 & 59.07 & 161.02 \\
\textbf{Candidate 2} & \textbf{56} & \textbf{4480} & \textbf{15680}  & \textbf{128} & \textbf{8.89}& \textbf{63.02} & \textbf{156.42 }\\
Candidate 3 & 56 & 4800 & 14400  & 120 & 8.97 & 62.94 & 155.86 \\ 
\bottomrule
\end{tabular}
\caption{Top 3 candidates for architecture selection. Accuracy is the average across reasoning benchmarks after distillation with 19B tokens. The last column shows vLLM output generation throughput (ISL/OSL=8k/16k and batch size=8).}
\label{tab:top3}
\end{table*}

\paragraph{FFN vs. Mamba pruning.}
We ablate the number of Mamba heads following the recipe in~\cite{taghibakhshi2025efficient}, considering configurations with 87.5\% and 93.75\% of the original heads. However, due to the relatively smaller compression ratios explored in this work (less than 15\% after depth pruning) compared to those in \cite{taghibakhshi2025efficient} (around 50\%), we find that applying Mamba head pruning yields limited benefit, and in these cases, pruning only the FFN and embedding dimensions—after depth pruning—proves sufficient to achieve the desired compression while preserving accuracy. Candidates 1 and 2 in Table~\ref{tab:top3} highlight this difference.

\subsection{Retraining with Distillation}
\label{sec:distillation}

To recover the accuracy lost due to pruning, the model undergoes continued training. Recent work has demonstrated that distilling knowledge from the original model to the pruned model outperforms conventional fine-tuning~\citep{muralidharan2024compactlanguagemodelspruning, sreenivas2024llmpruningdistillationpractice, bercovich2024puzzledistillationbasednasinferenceoptimized};
we thus adopt logit-based distillation for continued training, employing forward KL divergence loss exclusively during the accuracy recovery phase (see \S3 of the Minitron paper~\citep{muralidharan2024compactlanguagemodelspruning} for more details on the distillation loss formulation).
Building on the candidate selection process described in \S\ref{sec:nas}, we continue training Candidate 2 in an extended phase, as detailed below, to yield the final Nano 2 reasoning and base models.

\begin{table*}[!hbt]\small\centering
\setlength{\tabcolsep}{5pt}
\begin{tabular}{c|c|c}
\toprule
\textbf{\% Reasoning-SFT data} & \textbf{\% Pretraining data} &  \textbf{Accuracy (Avg)} \\ 
\midrule
50 & 50 & 57.5 \\
70 & 30 & \textbf{58.5} \\
90 & 10 & 57.2 \\ 
\bottomrule
\end{tabular}
\caption{Effect of varying reasoning data proportion on math accuracy after $\sim$ 6B tokens of KD.}
\label{tab:reasoning_data_ablation}
\end{table*}

\paragraph{Reasoning model.}
The reasoning model is distilled in stages with increasing sequence lengths to strengthen extended reasoning and long-context capabilities; this is followed by targeted reinforcement learning (RL), preference optimization and model merging to retain desired behaviors and ensure robustness across diverse tasks. We now describe these various stages:
\begin{enumerate}
    \item Depth pruning to 56 layers; Knowledge Distillation (KD) with $\sim$60B tokens at 8{,}192 sequence length.
    \item Width pruning and KD with:
        \begin{itemize}
            \item $\sim$50B tokens at 8{,}192 sequence length.
            \item $\sim$25B tokens at 49{,}152 sequence length.
            \item $\sim$1B tokens at 262{,}144 sequence length.
        \end{itemize}
    \item Direct Preference Optimization (DPO).
    \item Group Relative Policy Optimization (GRPO).
    \item KD with $\sim$0.4B tokens at 262{,}144 sequence length to recover post-RL drops.
    \item RLHF for alignment with human preferences.
    \item Model merging between steps \texttt{5} and \texttt{6} via 0.5 linear interpolation.
\end{enumerate}

\begin{figure}[!th]
    \centering
    \includegraphics[width=\linewidth]{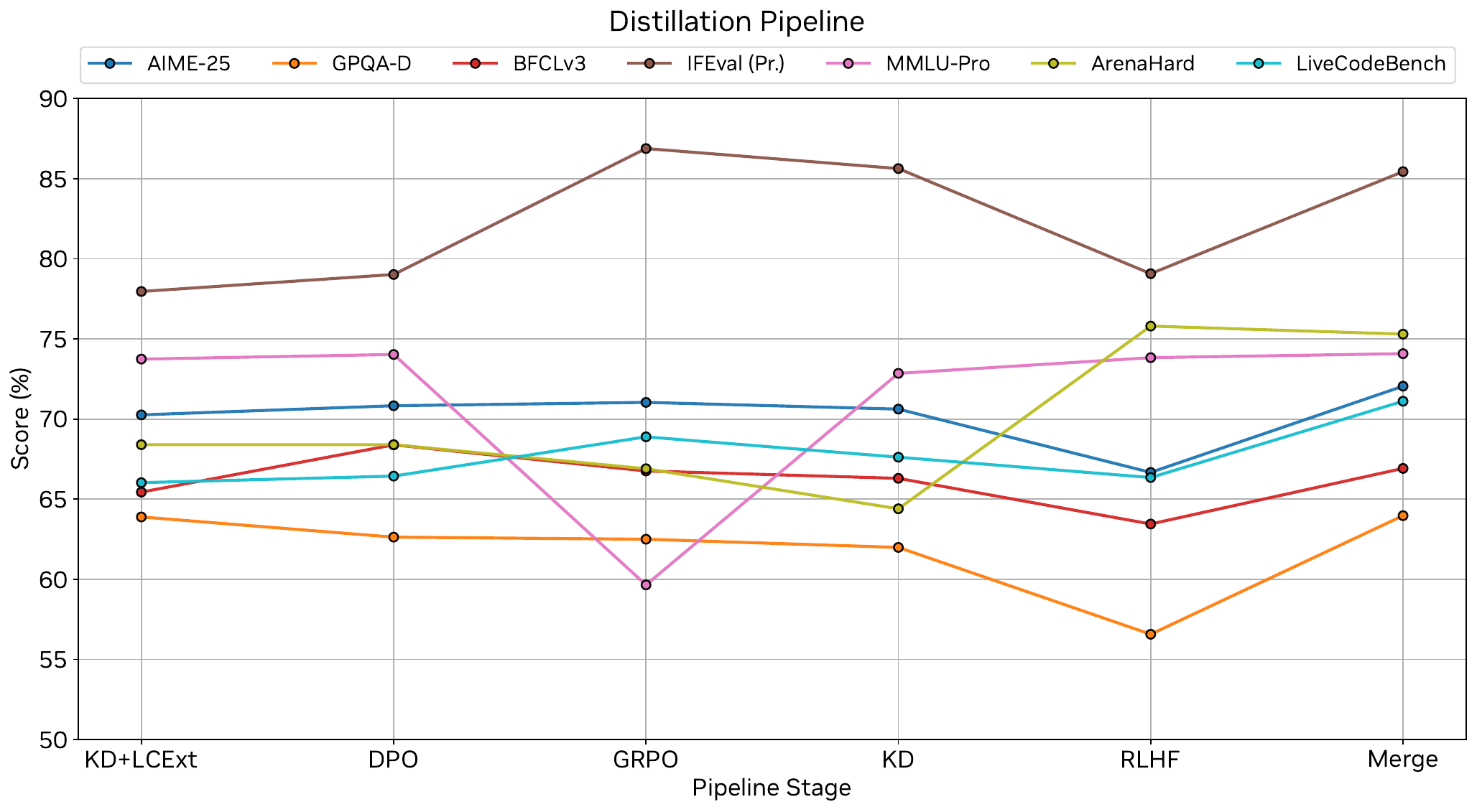}
    \caption{Task accuracy at different stages of the distillation pipeline for \ourmodel.}
    \label{fig:journey}
\end{figure}

More details on DPO, GRPO and RLHF can be found in Section~\ref{sec:alignment}.
Figure~\ref{fig:journey} shows the effects of staged training on model accuracy across different reasoning benchmarks.
Here, the $x$-axis represents the various stages (starting from Step 2 above), and the $y$-axis shows the scores obtained for the various benchmarks as training progresses.
As shown in the Figure, DPO and GRPO are critical for enhancing function-calling (BFCL v3) and instruction-following (IFEval) capabilities, though the latter temporarily degrades multi-task understanding (MMLU-Pro), which is recovered in the next step (post-GRPO KD). Finally, RLHF enhances alignment with human preferences (Arena-Hard) but causes additional benchmark drops, which are then recovered through model merging.

\textbf{Dataset:} We observe that a mix of 70\% post-training stage 2 data (Section~\ref{sec:supervised_fine_tuning}) and 30\% pretraining (Section~\ref{sec:pretrain_data}) data yields the highest accuracy (Table~\ref{tab:reasoning_data_ablation}). For KD at sequence length 262{,}144, we use 100\% stage 3 post-training data (Section~\ref{sec:supervised_fine_tuning}).

\paragraph{Base model.}
Distillation proceeds in stages: depth-only pruning and KD on $\sim$120B tokens, followed by width pruning and KD on $\sim$360B tokens (both at sequence length 8{,}192), and finally KD on $\sim$2.5B tokens at sequence length 524{,}288 to instill long-context capabilities.

\textbf{Dataset:} Following \cite{sreenivas2024llmpruningdistillationpractice}, we use 100\% pretraining data described in sections~\ref{sec:pretrain_data} and ~\ref{sec:pre-train-long-context} for distillation of the base model at sequence lengths 8{,}192 and 524{,}288, respectively.

\subsection{Results}

We efficiently compress the 12B model to 9B parameters by pruning full layers (depth), FFN hidden size, and embedding channels, improving inference throughput and enabling long-context inference on an NVIDIA A10G GPU. \ourfinalmodel retains 56 layers of the original model. Additionally, the number of embedding channels were pruned from 5120 to 4480, and FFN intermediate size was pruned from 20480 to 15680. As shown in Figure~\ref{fig:intro} and Tables~\ref{table:base_evals} and~\ref{table:base_evals_multilingual}, \ourfinalmodel achieves 3$\times$-6$\times$ higher throughput than Qwen3-8B for generation-heavy scenarios, while surpassing it in accuracy and remaining comparable to the 12B teacher on most benchmarks.
\section{Conclusion}
\label{sec:conclusion}
In this report, we introduced \ourfinalmodel, a hybrid Mamba-Transformer reasoning model that achieves comparable or better accuracies at up to 6$\times$ higher throughput than existing state-of-the-art models such as Qwen3-8B. To create \ourfinalmodel, we started by pre-training \ourbasemodel on 20T tokens, using a carefully constructed mix of curated and synthetically generated data. We aligned \ourbasemodel using several stages of SFT, GRPO, DPO, and RLHF before using the Minitron compression via pruning and distillation strategy to produce the final model. As a result of this compression, \ourfinalmodel can run inference on context lengths of up to 128k tokens in \texttt{bfloat16} precision on a single NVIDIA A10G GPU with 22 GiB of memory. We have open-sourced \ourfinalmodel along with its corresponding sibling \ourprunedbasemodel and parent \ourbasemodel models, plus the majority of its pre- and post-training data on HuggingFace (links at the bottom of Section~\ref{sec:intro}).

\section*{Contributors}

We thank the following people for their invaluable contributions to \ourmodelfull.

\textbf{Data.} Abhinav Khattar, Aleksander Ficek, Arham Mehta, Ayush Dattagupta, Brandon Norick, Dan Su, Daria Gitman, Evelina Bakhturina, Igor Gitman, Ivan Moshkov, Jaehun Jung, Jane Polak Scowcroft, Jocelyn Huang, Joseph Jennings, Jupinder Parmar, Markus Kliegl, Matvei Novikov, Mehrzad Samadi, Miguel Martinez, Mohammad Shoeybi, Mostofa Patwary, Pavlo Molchanov, Pritam Gundecha, Rabeeh Karimi Mahabadi, Ranjit Rajan, Rima Shahbazyan, Sanjeev Satheesh, Sarah Yurick, Sean Narenthiran, Seungju Han, Shizhe Diao, Shrimai Prabhumoye, Shubham Toshniwal, Siddhartha Jain, Somshubra Majumdar, Syeda Nahida Akter, Vahid Noroozi, Vineeth Kalluru, Vitaly Kurin, Wasi Uddin Ahmad, Wei Du, Ximing Lu, Yejin Choi, Ying Lin.

\textbf{FP8.} Hua Huang, Jinze Xue, Keith Wyss, Kunlun Li, Mike Chrzanowski, Oleg Rybakov, Przemek Tredak, Tim Moon, Zhongbo Zhu.

\textbf{Architecture.} Bita Darvish Rouhani, Brandon Norick, Duncan Riach, Nidhi Bhatia, Roger Waleffe, Wonmin Byeon, Ritika Borkar, Xin Dong, Yonggan Fu.

\textbf{Pretraining.} Aarti Basant, Abhijit Paithankar, Abhinav Khattar, Deepak Narayanan, Herman Sahota, Hexin Wang, Jupinder Parmar, Mohammad Shoeybi, Mostofa Patwary, Namit Dhameja, Roger Waleffe, Russell J. Hewett, Ryan Prenger, Seonmyeong Bak.

\textbf{Infrastructure.} Alex Kondratenko, Alex Shaposhnikov, Anubhav Mandarwal, Ashwin Poojary, Dong Ahn, Gargi Prasad, Haim Elisha, Harsh Sharma, Kumar Anik, Maer Rodrigues de Melo, Ruoxi Zhang, Shelby Thomas, Stefania Alborghetti, Tony Wang.

\textbf{Long Context.} Deepak Narayanan, Dima Rekesh, Duncan Riach, John Kamalu, Kezhi Kong, Markus Kliegl, Roger Waleffe, Samuel Kriman.

\textbf{Inference.} Daniel Afrimi, Helen Ngo, Keshav Santhanam, Kushan Ahmadian, Lawrence McAfee, Luis Vega, Nave Assaf, Peter Dykas, Shanmugam Ramasamy, Siddharth Singh, Tomer Asida, Vijay Korthikanti.

\textbf{Alignment.} Adithya Renduchintala, Alexander Bukharin, Ameya Sunil Mahabaleshwarkar, Banghua Zhu, Bilal Kartal, Brian Yu, Charles Wang, Christian Munley, David Mosallanezhad, Gerald Shen, Haifeng Qian, Hayley Ross, Hoo Chang Shin, Igor Gitman, Jian Zhang, Jiaqi Zeng, Julien Veron Vialard, Junkeun Yi, Kezhi Kong, Luis Vega, Makesh Narsimhan Sreedhar, Oleksii Hrinchuk, Oleksii Kuchaiev, Peter Jin, Prasoon Varshney, Ritu Gala, Shuoyang Ding, Soumye Singhal, Tugrul Konuk, Venkat Srinivasan, Vitaly Lavrukhin, Yian Zhang, Yoshi Suhara, Zhen Dong, Zijia Chen.

\textbf{Compression.} Aditya Malte, Akhiad Bercovich, Akshay Hazare, Ali Taghibakhshi, Ameya Sunil Mahabaleshwarkar, Ashwath Aithal, Banghua Zhu, Daniel Korzekwa, Deepak Narayanan, Gerald Shen, Hayley Ross, Julien Veron Vialard, Luis Vega, Marcin Chochowski, Mostofa Patwary, Nima Tajbakhsh, Oluwatobi Olabiyi, Pavlo Molchanov, Ran El-Yaniv, Roger Waleffe, Saurav Muralidharan, Sepehr Sameni, Sharath Turuvekere Sreenivas, Tomer Asida, Yashaswi Karnati, Yian Zhang, Yoshi Suhara, Zijia Chen.

\textbf{Software Support.} Abhijit Khairnar, Adithya Renduchintala, Ali Taghibakhshi, Anna Shors, Ashwath Aithal, Balaram Buddharaju, Bobby Chen, Charlie Truong, Deepak Narayanan, Dmytro Pykhtar, Duncan Riach, Gerald Shen, Helen Ngo, Jared Casper, Jimmy Zhang, Keshav Santhanam, Kezhi Kong, Lawrence McAfee, Luis Vega, Nima Tajbakhsh, Parth Chadha, Piotr Bialecki, Prashant Gaikwad, Rajen Patel, Roger Waleffe, Sahil Jain, Terry Kong, Tyler Poon, Vijay Korthikanti, Vikram Fugro, Yoshi Suhara, Zhiyu Li.

\textbf{Evaluations and Safety.} Christopher Parisien, Dan Su, Daniel Rohrer, Eileen Long, Erick Galinkin, Helen Ngo, Katherine Luna, Keshav Santhanam, Kezhi Kong, Leon Derczynski, Marta Stepniewska-Dziubinska, Meriem Boubdir, Michal Bien, Michael Boone, Michael Evans, Michal Bien, Michal Zawalski, Pablo Ribalta, Piotr Januszewski, Pradeep Thalasta, Sanjeev Satheesh, Shaona Ghosh, Tomasz Hliwiak. 

\textbf{Legal and Compliance.} Barnaby Simkin, Chetan Mungekar, Dina Yared, Iain Cunningham, Katherine Cheung, Laya Sleiman, Meredith Price, Michael Boone, Nikki Pope, Ria Cheruvu, Saori Kaji.

\textbf{Marketing.} Amelia Barton, Chris Alexiuk, Mark Cai, Nirmal Kumar Juluru, Shreya Gopal.

\textbf{Project Management.} Alejandra Rico, Amy Shen, Ann Guan, Ashton Sharabiani, Elliott Ning, Krzysztof Pawelec, Negar Habibi, Twinkle Vashishth.

\textbf{Product.} Arun Venkatesan, Chintan Patel, Chris Alexiuk, Joey Conway, Padmavathy Subramanian, Udi Karpas.

\textbf{Leadership.} Andrew Tao, Boris Ginsburg, Bryan Catanzaro, Eric Chung, Jan Kautz, Joey Conway, Jonathan Cohen, Kari Briski, Mohammad Shoeybi, Mostofa Patwary, Oleksii Kuchaiev, Pavlo Molchanov.

\textit{We also thank Chen Zhang, Michael Goin, Thomas Parnell from the vLLM team for their assistance.}

\newpage

\bibliography{references}
\bibliographystyle{references}

\appendix

\section{Permissive Source Code Licenses}
\label{appendix:acceptable-licenses}

We remove source code with a license not in the following list:

\begin{simplechar}
3Com Microcode 3com-microcode, 3D Slicer License 1.0 [3dslicer-1.0], 4Suite 1.1 [4suite-1.1], AAL [attribution], Abstyles License [abstyles], ACE TAO License [ace-tao], AdaCore Doc License [adacore-doc], ADI BSD [adi-bsd], Adobe Glyph License [adobe-glyph], Adobe Postscript AFM License [apafml], Adobe Source Code License 2006 [adobe-scl], AES-128 3.0 License [aes-128-3.0], AFL 1.1 [afl-1.1], AFL 1.2 [afl-1.2], AFL 2.0 [afl-2.0], AFL 2.1 [afl-2.1], AFL 3.0 [afl-3.0], afmparse License [afmparse], Agere BSD [agere-bsd], Alexisisaac Freeware License [alexisisaac-freeware], Allegro 4 License [allegro-4], Altera License [xnet], Amazon Digital Services License [adsl], AMD Historical License [amd-historical], AMD PLPA License [amdplpa], AMPAS BSD-Style License [ampas], AMSFonts license [ams-fonts], Andre Adrian DFS license [adrian], ANTLR-PD [antlr-pd], ANTLR-PD with fallback [antlr-pd-fallback], ANU License [anu-license], Apache 1.0 [apache-1.0], Apache 1.1 [apache-1.1], Apache 2.0 [apache-2.0], Apache Patent Provision Exception Terms [apache-patent-exception], App::s2p License [app-s2p], Apple Attribution 1997 [apple-attribution-1997], Apple Attribution License [apple-attribution], Apple Example Code License [apple-excl], Apple MIT License [aml], Apple Sample Source Code License [apple-sscl], Aravindan Premkumar Licenase [aravindan-premkumar], ArgoUML License [argouml], ARM LLVM Grant [arm-llvm-sga], Array Input Method Public License [array-input-method-pl], Artistic 1.0 [artistic-1.0], Artistic 1.0 w/clause 8 [artistic-1.0-cl8], Artistic 2.0 [artistic-2.0], Artistic-Perl-1.0 [artistic-perl-1.0], ASMUS License [asmus], ASN.1 Object Dumping Code License [asn1], Atkinson Hyperlegible Font License [atkinson-hyperlegible-font], Baekmuk Fonts License [baekmuk-fonts], Bahyph License [bahyph], BaKoMa Fonts Licence 1995 [bakoma-fonts-1995], Barr TeX License [barr-tex], BEA 2.1 [bea-2.1], Beal Screamer License [beal-screamer], Beer-Ware License [beerware], BERI Hardware-Software License v1.0 [beri-hw-sw-1.0], BigDigits License [bigdigits], Bigelow & Holmes Lucida Fonts License [bigelow-holmes], Biopython License [biopython], Bitstream Vera Font License [bitstream], Bitzi-PD [bitzi-pd], BLAS License 2017 [blas-2017], Blue Oak Model License 1.0.0 [blueoak-1.0.0], BOHL-0.2 [bohl-0.2], Boost 1.0 [boost-1.0], Boost Original [boost-original], Borceux License [borceux], Boutell libgd declarations 2021 [boutell-libgd-2021], bpmn.io License [bpmn-io], Brent Corkum License [brent-corkum], Brian Clapper License [brian-clapper], Brian Gladman 3-Clause License [brian-gladman-3-clause], Brian Gladman Dual BSD-GPL [brian-gladman-dual], Brian Gladman License [brian-gladman], Broadcom CFE License [broadcom-cfe], Broadcom Warranty Disclaimer [broadcom-linux-timer], Brocade Firmware License [brocade-firmware], Bruno Podetti License [bruno-podetti], BSD 1988 [bsd-1988], BSD 3-Clause Devine [bsd-3-clause-devine], BSD 3-Clause FDA [bsd-3-clause-fda], BSD 3-Clause jtag [bsd-3-clause-jtag], BSD 3-Clause No Change [bsd-3-clause-no-change], BSD 3-Clause No Nuclear Warranty [bsd-3-clause-no-nuclear-warranty], BSD 3-Clause no trademark [bsd-3-clause-no-trademark], BSD 3-Clause Open MPI variant [bsd-3-clause-open-mpi], BSD 3-Clause Sun [bsd-3-clause-sun], BSD 3-Clause with GPL reference [bsd-top-gpl-addition], BSD Acknowledgment (Carrot2) License [bsd-ack-carrot2], BSD Acknowledgment License [bsd-ack], BSD Advertising Acknowledgement License [bsd-advertising-acknowledgement], BSD Artwork [bsd-artwork], BSD Atmel License [bsd-atmel], BSD DPT [bsd-dpt], BSD plus modification notice [bsd-plus-mod-notice], BSD Simplified Darwin [bsd-simplified-darwin], BSD Source Code Attribution [bsd-source-code], BSD Unchanged [bsd-unchanged], BSD Unmodified [bsd-unmodified], BSD Zero Clause License [bsd-zero], BSD-1-Clause [bsd-1-clause], BSD-1-Clause Build [bsd-1-clause-build], BSD-2-Clause [bsd-simplified], BSD-2-Clause no disclaimer [bsd-no-disclaimer], BSD-2-Clause no disclaimer Unmod [bsd-no-disclaimer-unmodified], BSD-2-Clause Plus Patent [bsd-plus-patent], BSD-2-Clause-plus-advertizing [bsd-2-clause-plus-advertizing], BSD-2-Clause-Views [bsd-2-clause-views], BSD-3-Clause [bsd-new], BSD-3-Clause tcpdump variant [bsd-new-tcpdump], BSD-3-Clause without notice modification [bsd-new-nomod], BSD-3-Clause X11 disclaimer [bsd-x11], BSD-4-Clause with Voices [bsd-original-voices], BSD-4-Clause-Shortened [bsd-4-clause-shortened], BSD-Axis without modification [bsd-axis-nomod], BSD-Credit [bsd-credit], BSD-Derivative [bsd-new-derivative], BSD-Export [bsd-export], BSD-InnoSys [bsd-innosys], BSD-Mylex [bsd-mylex], BSD-Original [bsd-original], BSD-Original-Muscle [bsd-original-muscle], BSD-Original-UC [bsd-original-uc], BSD-Original-UC-1986 [bsd-original-uc-1986], BSD-Simplified Intel [bsd-simplified-intel], BSD-Simplified source [bsd-simplified-source], BSD-Top [bsd-top], BSLA [bsla], BSLA no advertizing [bsla-no-advert], Business Source License 1.0 [bsl-1.0], BYTEmark License [bytemark], bzip2 License 2010 [bzip2-libbzip-2010], Caldera License [caldera], Careware [careware], Carnegie Mellon Contributors [carnegie-mellon-contributors], Carnegie Mellon License [carnegie-mellon], Cavium malloc License [cavium-malloc], CC-BY-1.0 [cc-by-1.0], CC-BY-2.0 [cc-by-2.0], CC-BY-2.0-UK [cc-by-2.0-uk], CC-BY-2.5 [cc-by-2.5], CC-BY-3.0 [cc-by-3.0], CC-BY-3.0-AT [cc-by-3.0-at], CC-BY-3.0-US [cc-by-3.0-us], CC-BY-4.0 [cc-by-4.0], CC-PD [cc-pd], CC-PD Mark 1.0 [cc-pdm-1.0], CC0-1.0 [cc0-1.0], CDLA Permissive 1.0 [cdla-permissive-1.0], CDLA Permissive 2.0 [cdla-permissive-2.0], CeCILL-B License [cecill-b], CeCILL-B License English [cecill-b-en], CERN Attribution 1995 [cern-attribution-1995], CERN Open Hardware Licence v1.2 [cern-ohl-1.2], CERN Open Hardware License v1.1 [cern-ohl-1.1], CERN-OHL-P-2.0 [cern-ohl-p-2.0], CFITSIO License [cfitsio], Checkmk License [checkmk], Chicken Dance License v0.2 [chicken-dl-0.2], Chris Maunder License [chris-maunder], Chris Stoy Attribution License [chris-stoy], Clarified Artistic License [artistic-clarified], Classic VB License [classic-vb], Clear BSD 1-Clause License [clear-bsd-1-clause], Clear BSD License [clear-bsd], Click License [click-license], CLIPS License 2017 [clips-2017], CMU Computing Services License [cmu-computing-services], CMU License [cmu-template], CMU MIT-style [cmu-mit], CMU Simple License [cmu-simple], CMU Style [cmu-uc], CNRI Jython License [cnri-jython], CNRI Python 1.6 [cnri-python-1.6], CNRI Python 1.6.1 [cnri-python-1.6.1], Code Credit License v1.0.1 [code-credit-license-1.0.1], Code Credit License v1.1.0 [code-credit-license-1.1.0], CodeGuru Permissions [codeguru-permissions], CodeSourcery 2004 [codesourcery-2004], COIL-1.0 [coil-1.0], Common Lisp LOOP License [loop], CommonJ Timer License [commonj-timer], Compass License [compass], ComponentAce JCraft License [componentace-jcraft], compuphase Linking Exception to Apache 2.0 [compuphase-linking-exception], Condor Public License 1.1 [condor-1.1], Copyheart [copyheart], Cornell Lossless JPEG License [cornell-lossless-jpeg], Cougaar Open Source License [cosl], CP/M License 2022 [cpm-2022], CppCoreGuidelines License [cpp-core-guidelines], CRCalc license [crcalc], Creative Commons Attribution 2.5 Australia [cc-by-2.5-au], Creative Commons Attribution 3.0 Germany [cc-by-3.0-de], Creative Commons Attribution 3.0 Netherlands [cc-by-3.0-nl], Crossword License [crossword], Crypto++ License [cryptopp], Crystal Stacker License [crystal-stacker], CSL-1.0 [csl-1.0], CSPRNG [csprng], Cube License [cube], cURL License [curl], CVE ToU [cve-tou], CWE ToU [cwe-tou], CxImage License [cximage], D Zlib [d-zlib], DAMAIL [damail], Dante Treglia License [dante-treglia], DBAD License 1.1 [dbad-1.1], Debian reportbug License [reportbug], Delorie Historical License [delorie-historical], dhtmlab Public License [dhtmlab-public], diffmark License [diffmark], dl-de/by-1-0-de [dl-de-by-1-0-de], dl-de/by-1-0-en [dl-de-by-1-0-en], dl-de/by-2-0-de [dl-de-by-2-0-de], dl-de/by-2-0-en [dl-de-by-2-0-en], dmalloc License [dmalloc], DMTF License 2017 [dmtf-2017], Docbook License [docbook], Dom4j License [dom4j], Dotseqn License [dotseqn], Douglas Young License [douglas-young], DRL-1.0 [drl-1.0], DRL-1.1 [drl-1.1], Dropbear License [dropbear], Dropbear-2016 [dropbear-2016], DSDP License [dsdp], Dtree License [dtree], dvipdfm License [dvipdfm], DWTFNMFPL-3.0 [dwtfnmfpl-3.0], Dynamic Drive TOU [dynamic-drive-tou], ECL 1.0 [ecl-1.0], ECL 2.0 [ecl-2.0], EFL 1.0 [efl-1.0], EFL 2.0 [efl-2.0], EFL MIT-Style License [enlightenment], eGenix Public License 1.0.0 [egenix-1.0.0], eGenix Public License 1.1.0 [egenix-1.1.0], EllisLab License [ellis-lab], EMX Library License [emx-library], EnergyPlus BSD-Style License [energyplus-bsd], Enhanced MIT License [emit], enna License [enna], Entessa 1.0 [entessa-1.0], ePaperPress License [epaperpress], EPICS Open License [epics], Eric Glass License [eric-glass], Errbot exception [errbot-exception], Etalab Open License 2.0 [etalab-2.0], Etalab Open License 2.0 English [etalab-2.0-en], EU DataGrid Software License [eu-datagrid], Fabien Tassin License [fabien-tassin], Fair License [fair], FAL 1.3 [free-art-1.3], Far Manager exception to BSD-3-Clause [far-manager-exception], FASTBuild License 2012-2020 [fastbuild-2012-2020], FastCGI DevKit [fastcgi-devkit], FastCGI License for Spec Implementation [openmarket-fastcgi], FatFs License [fatfs], FFTPACK License 2004 [fftpack-2004], Filament Group MIT License [filament-group-mit], Flex 2.5 [flex-2.5], Flora License v1.1 [flora-1.1], font-alias License [font-alias], FPLOT LIcense [fplot], Fraunhofer ISO 14496-10 License [fraunhofer-iso-14496-10], FreeBSD Boot [freebsd-boot], FreeBSD Doc License [freebsd-doc], FreeBSD unmodified first lines License [freebsd-first], FreeMarker License [freemarker], FreeTTS License [freetts], FreeType Project License [freetype], Freeware Public License (FPL) [fpl], FSF All Permissive License [fsf-ap], FSF Free Software License [fsf-free], FSF Notice [fsf-notice], FSF Unlimited License No Warranty [fsf-unlimited-no-warranty], FSF-Unlimited [fsf-unlimited], Fujion Clinical Exception to Apache 2.0 [fujion-exception-to-apache-2.0], Gareth McCaughan License [gareth-mccaughan], Gary S. Brown License [gary-s-brown], GDCL License [gdcl], Generic patent disclaimer [patent-disclaimer], Geoff Kuenning License 1993 [geoff-kuenning-1993], Ghostpdl Permissive [ghostpdl-permissive], Glulxe License [glulxe], GLUT License [glut], GLWTPL [glwtpl], Good Boy License [good-boy], Graphics Gems License [graphics-gems], Greg Roelofs License [greg-roelofs], Gregory Pietsch Liberal License [gregory-pietsch], GStreamer Exception (2005) [gstreamer-exception-2005], GTPL-v1 [gtpl-v1], GTPL-v2 [gtpl-v2], GTPL-v3 [gtpl-v3], Haskell Report License [haskell-report], HDF4 License [hdf4], HDF5License [hdf5], HDPARM License [hdparm], Henry Spencer License 1999 [henry-spencer-1999], Henry Spencer Regexp License [hs-regexp], HIDAPI License [hidapi], Historical Notice - NTP [historical-ntp], Historical Permission Notice and Disclaimer [historical], Homebrewed License [homebrewed], HP 1986 License [hp-1986], HPND sell variant with MIT disclaimer [hpnd-sell-variant-mit-disclaimer], HTML 5 spec License [html5], httpget notice and disclaimer [httpget], Ian Kaplan License [ian-kaplan], Ian Piumarta License [ian-piumarta], IBM AS-IS License [ibm-as-is], IBM DHCP License [ibm-dhcp], IBM Non-Warranted Sample Code License [ibm-nwsc], IBM PowerPC Software [ibm-pibs], IBM Sample License [ibm-sample], IBPP License [ibpp], ICANN-Public [icann-public], ICOT Free Software [icot-free], ICU Composite License [ibm-icu], ICU License 58 and later [unicode-icu-58], IDT License Notice [idt-notice], IETF License [ietf], IETF Trust License [ietf-trust], ilmid License [ilmid], ImageMagick License [imagemagick], Independent JPEG Group License - short [ijg-short], Indiana Extreme License 1.1.1 [indiana-extreme], Indiana Extreme License 1.2 [indiana-extreme-1.2], Infineon Free Software License [infineon-free], Info-Zip License 1997-10 [info-zip-1997-10], Info-Zip License 2001-01 [info-zip-2001-01], Info-Zip License 2002-02 [info-zip-2002-02], Info-Zip License 2003-05 [info-zip-2003-05], Info-Zip License 2004-05 [info-zip-2004-05], Info-Zip License 2005-02 [info-zip-2005-02], Info-Zip License 2007-03 [info-zip-2007-03], Info-Zip License 2009-01 [info-zip-2009-01], Info-Zip License [info-zip], Inno Setup License [inno-setup], Intel ACPI SLA [intel-acpi], Intel BSD - Export Control [intel-bsd-export-control], Intel BSD 2 Clause License [intel-bsd-2-clause], Intel BSD License [intel-bsd], Intel Limited Patent License [intel], Intel OSL 1989 [intel-osl-1989], Intel OSL 1993 [intel-osl-1993], Intel Royalty Free License [intel-royalty-free], ISC License [isc], ISO 14496-10 [iso-14496-10], ISO 8879 [iso-8879], ITU License [itu], JA-SiG License [ja-sig], Jam License [jam], Jason Mayes License [jason-mayes], Jasper 1.0 [jasper-1.0], JasPer 2.0 [jasper-2.0], Java App Stub License [java-app-stub], JDBM License v1.00 [jdbm-1.00], JDOM License [jdom], Jetty License [jetty], JGraph License [jgraph], JPEG License [ijg], JPNIC idnkit License [jpnic-idnkit], JPNIC mdnkit License [jpnic-mdnkit], JPython 1.1 [jpython-1.1], jQuery-Tools-PD [jquery-pd], Jscheme License [jscheme], JSFromHell License [jsfromhell], JSON License [json], JSON-js-PD [json-js-pd], JSON-PD [json-pd], Jython License [jython], Kalle Kaukonen License [kalle-kaukonen], Kazlib [kazlib], Keith Rule License [keith-rule], Kerberos License [kerberos], Kevan Stannard License [kevan-stannard], Kevlin Henney License [kevlin-henney], Khronos License [khronos], Knuth CTAN License [knuth-ctan], Kumar Robotics License [kumar-robotics], latex-ec-fonts [ecfonts-1.0], Latex2e License [latex2e], Latex2e with translated notice permission [latex2e-translated-notice], LBNL BSD Variant [lbnl-bsd], LCS-Telegraphics License [lcs-telegraphics], Leptonica License [leptonica], libgd License 2018 [libgd-2018], libgeoTiff License [libgeotiff], LibMib License [libmib], libmng License 2007 [libmng-2007], Libpng License [libpng], LIbpng License v2 [libpng-v2], libselinux License [libselinux-pd], libsrv License v1.0.2 [libsrv-1.0.2], Lil License v1 [lil-1], LILO License [lilo], Linux Device Drivers [linux-device-drivers], Linux-OpenIB [linux-openib], LinuxBIOS License [linuxbios], linuxhowtos License [linuxhowtos], LLNL [llnl], LLVM Exception to Apache 2.0 [llvm-exception], Logica OSL 1.0 [logica-1.0], LPPL 1.3c [lppl-1.3c], Lucent Public License 1.0 [lucent-pl-1.0], Lucent Public License 1.02 [lucent-pl-1.02], Lucre License [lucre], LZMA SDK License (versions 9.22 and beyond) [lzma-sdk-9.22], LZMA SDK Public Domain [lzma-sdk-pd], M+ Fonts license [m-plus], MakeHuman License [make-human-exception], Markus Kuhn License [markus-kuhn-license], Martin Bergmeier License [martin-birgmeier], Matrix Template Library License [mtll], Matt Gallagher Attribution License [matt-gallagher-attribution], Matt Kruse License [mattkruse], Matthew Kwan License [matthew-kwan], MediaInfo(Lib) License [mediainfo-lib], metamail License [metamail], MgOpen Font License [mgopen-font-license], Michael Barr License [michael-barr], Minpack Copyright Notice [minpack], MirOS License [mir-os], MIT (SEI) [vince], MIT 1995 [mit-1995], MIT Acknowledgment License [mit-ack], MIT Addition License [mit-addition], MIT License 1998 [mit-license-1998], MIT License [mit], MIT Modern Variant [mit-modern], MIT Nagy Variant [mit-nagy], MIT no advertising with Export Control [mit-no-advert-export-control], MIT No Commercial Use of Trademarks [mit-no-trademarks], MIT no false attribution License [mit-no-false-attribs], MIT Old Style [mit-old-style], MIT Old Style no advertising [mit-old-style-no-advert], MIT Old Style Spare [mit-old-style-sparse], MIT README License [mit-readme], MIT Synopsys License [mit-synopsys], MIT Taylor Variant [mit-taylor-variant], MIT Veillard Variant [mit-veillard-variant], MIT with Export Control [mit-export-control], MIT with Specification Disclaimer [mit-specification-disclaimer], MIT Xfig Variant [mit-xfig], MIT-0-Clause [mit-0], mod_dav License 1.0 [mod-dav-1.0], Modified MIT License for Public Domain software [pd-mit], Motorola Microprocessor License [motorola], Mozilla GC License [mozilla-gc], MPEG SSG License [mpeg-ssg], MPEG-2 NBC MPEG-4 Audio ISO [mpeg-iso], MPICH License [mpich], MS Systems Journal Sample Code License [msj-sample-code], MS WS Routing Specifications License [ms-ws-routing-spec], MS-LPL [ms-lpl], MS-PL [ms-pl], MS-SS-PL [ms-sspl], Mulan PSL v1 [mulanpsl-1.0], Mulan PSL v1.0 (En) [mulanpsl-1.0-en], Mulan PSL v2 [mulanpsl-2.0], Mulan PSL v2.0 (En) [mulanpsl-2.0-en], Mulle Kybernetik License [mulle-kybernetik], Multics License [multics], Mup License [mup], musl attribution exception [musl-exception], MX4J License 1.0 [mx4j], Nara Institute License 2003 [naist-2003], NASA 1.3 [nasa-1.3], NAUMEN Public License [naumen], NBPL-1.0 [nbpl-1.0], NCBI Public Domain Notice [ncbi], NCSA Open Source License [uoi-ncsa], Net SNMP License [net-snmp], Netcat License [netcat], NetCDF License [netcdf], Netron Project License [netron], Newlib Historical License [newlib-historical], Newran License [newran], Newsletr License [newsletr], Nice License [nice], NICTA Public Software Licence 1.0 [nicta-psl], Niels Ferguson License [niels-ferguson], Nilsson Historical License [nilsson-historical], NIST Public Domain Notice [nist-pd], NIST Public Domain Notice with fallback [nist-pd-fallback], NIST Software License [nist-software], NIST SRD License [nist-srd], NLOD-1.0 [nlod-1.0], NLOD-2.0 [nlod-2.0], NLPL [nlpl], Node License [node-js], Non White Heterosexual Male [nwhm], Nonexclusive License [nonexclusive], Nortel DASA License [nortel-dasa], Notre Dame License [notre-dame], NRL License [nrl], NRL permission [nrl-permission], NTLM License [ntlm], NTP Origin License [ntpl-origin], NTP-0 [ntp-0], NVIDIA 2002 License [nvidia-2002], NVIDIA License [nvidia], NVIDIA License with Government Qualifications [nvidia-gov], NYSL 0.9982 [nysl-0.9982], NYSL 0.9982 JP [nysl-0.9982-jp], O Young Jong License [o-young-jong], O'Reilly Code Sample Notice [oreilly-notice], O-UDA-1.0 [o-uda-1.0], Oasis WS Security Specification License [oasis-ws-security-spec], Object Form Exception to MIT [object-form-exception-to-mit], ODC-By-1.0 [odc-by-1.0], ODMG License [odmg], OFFIS License [offis], OFL 1.0 [ofl-1.0], OFL 1.0 no Reserved Font Name [ofl-1.0-no-rfn], OFL 1.0 Reserved Font Name [ofl-1.0-rfn], OFL 1.1 no Reserved Font Name [ofl-1.1-no-rfn], OGC 1.0 [ogc-1.0], OGC Software Notice [ogc], OGL 1.0a [ogl-1.0a], OGL Alberta 2.1 [can-ogl-alberta-2.1], OGL British Columbia 2.0 [can-ogl-british-columbia-2.0], OGL Canada 2.0 [can-ogl-2.0-en], OGL Canada 2.0 Francais [ogl-canada-2.0-fr], OGL Nova Scotia 1.0 [can-ogl-nova-scotia-1.0], OGL Ontario 1.0 [can-ogl-ontario-1.0], OGL Toronto 1.0 [can-ogl-toronto-1.0], OGL-UK-1.0 [ogl-uk-1.0], OGL-UK-2.0 [ogl-uk-2.0], OGL-UK-3.0 [ogl-uk-3.0], OGL-WPD-3.0 [ogl-wpd-3.0], Open Directory License [odl], Open Group Test Suite License [opengroup], Open Publication License 1.0 [openpub], OpenLDAP Public License 1.1 [openldap-1.1], OpenLDAP Public License 1.2 [openldap-1.2], OpenLDAP Public License 1.3 [openldap-1.3], OpenLDAP Public License 1.4 [openldap-1.4], OpenLDAP Public License 2.0 [openldap-2.0], OpenLDAP Public License 2.0.1 [openldap-2.0.1], OpenLDAP Public License 2.1 [openldap-2.1], OpenLDAP Public License 2.2 [openldap-2.2], OpenLDAP Public License 2.2.1 [openldap-2.2.1], OpenLDAP Public License 2.2.2 [openldap-2.2.2], OpenLDAP Public License 2.3 [openldap-2.3], OpenLDAP Public License 2.4 [openldap-2.4], OpenLDAP Public License 2.5 [openldap-2.5], OpenLDAP Public License 2.6 [openldap-2.6], OpenLDAP Public License 2.7 [openldap-2.7], OpenLDAP Public License 2.8 [openldap-2.8], OpenORB Community License 1.0 [openorb-1.0], OpenSAML License v1 [opensaml-1.0], OpenSSH License [openssh], OpenSSL License [openssl], OpenSSL/SSLeay License [openssl-ssleay], OPML 1.0 [opml-1.0], OPNL-1.0 [opnl-1.0], OPNL-2.0 [opnl-2.0], Oracle BSD-Style with Nuclear Restrictions [oracle-bsd-no-nuclear], Original SSLeay License [ssleay], Original SSLeay License with Windows Clause [ssleay-windows], Oswego Concurrent License [oswego-concurrent], Other Permissive Licenses [other-permissive], OWTChart License [owtchart], OZPLB 1.0 [ozplb-1.0], OZPLB 1.1 [ozplb-1.1], Paolo Messina 2000 [paolo-messina-2000], ParaView License 1.2 [paraview-1.2], Paul Mackerras Binary License [paul-mackerras-binary], Paul Mackerras License [paul-mackerras], Paul Mackerras New License [paul-mackerras-new], Paul Mackerras Simplified License [paul-mackerras-simplified], Paulo Soares License [paulo-soares], PayPal SDK License 2013-2016 [paypal-sdk-2013-2016], PBM Library License [libpbm], PCRE License [pcre], PD'Programming License [pd-programming], PDDL 1.0 [pddl-1.0], Perl 1.0 [perl-1.0], Peter Deutsch Document License [peter-deutsch-document], Phil Bunce License [phil-bunce], Philippe De Muyter License [philippe-de-muyter], Phorum License 2.0 [phorum-2.0], PHP License 2.0.2 [php-2.0.2], PHP License 3.0 [php-3.0], PHP License 3.01 [php-3.01], Pine License [pine], PngSuite License [pngsuite], Politepix Public License 1.0 [politepix-pl-1.0], PostgreSQL License [postgresql], ppp License [ppp], Protobuf License [protobuf], PS Utilities License [psutils], PSF Python License 3.7.2 [psf-3.7.2], PSF-2.0 [psf-2.0], psfrag License [psfrag], Psytec Free Software License [psytec-freesoft], Public Domain [public-domain], Public Domain Disclaimer [public-domain-disclaimer], Purdue BSD-Style License [purdue-bsd], pybench License [pybench], PyCrypto License [pycrypto], PyGres License 2.2 [pygres-2.2], Python CWI License [python-cwi], Python License 2.0 [python], Python License 2.0.1 [python-2.0.1], Qhull License [qhull], QLogic Microcode [qlogic-microcode], Qpopper License [qpopper], Qualcomm Turing License [qualcomm-turing], Quirksmode Copyright Notice [quirksmode], radvd License [radvd], Rdisc License [rdisc], Red Hat Attribution License [red-hat-attribution], Red Hat BSD-Simplified [red-hat-bsd-simplified], Regexp License [regexp], Repoze License [repoze], RiceBSD [ricebsd], Richard Black License [richard-black], Robert Hubley License [robert-hubley], RSA 1990 [rsa-1990], RSA Cryptoki License [rsa-cryptoki], RSA Demo License [rsa-demo], RSA-MD4 License [rsa-md4], RSA-MD5 License [rsa-md5], RTools.Util License [rtools-util], Ruby License [ruby], Runtime Library Exception to Apache 2.0 [apple-runtime-library-exception], Rute Users Tutorial and Exposition License 0.8.0 [rute], Ryszard Szopa License [ryszard-szopa], SaaS MIT License [saas-mit], Sash Notice [sash], SATA License [sata], SAX-PD [sax-pd], Saxpath License [saxpath], SBIA Part B [sbia-b], ScanCode acknowledgment [scancode-acknowledgment], scanlogd License [scanlogd-license], ScanSoft Public License 1.2 [scansoft-1.2], SCEA Shared Source License 1.0 [scea-1.0], Scheme Language Report License [schemereport], Scheme Widget Library (SWL) Software License [swl], Scintilla License [scintilla], Scribbles Demos Recognizer Notice [scribbles], Script Asylum License [script-asylum], Secret Labs License 2011 [secret-labs-2011], selinux-nsa-declaration-1.0 [selinux-nsa-declaration-1.0], Sendmail License [sendmail], Service Availability Forum License [saf], Service Component Architecture License [service-comp-arch], SFL License Agreement [sfl-license], SGI CID Font Code Public License 1.0 [sgi-cid-1.0], SGI Free Software License B 1.1 [sgi-freeb-1.1], SGI Free Software License B 2.0 [sgi-freeb-2.0], SGI GLX Public License 1.0 [sgi-glx-1.0], Sglib License [sglib], SGP4 Permission Notice [sgp4], Shital Shah License [shital-shah], SIL Open Font License 1.1 with Reserved Font Name [ofl-1.1-rfn], SimPL 1.1 [simpl-1.1], SNMP++ License [hp-snmp-pp], snprintf License [snprintf], SoftFloat [softfloat], SoftFloat Legal Notice 2.0 [softfloat-2.0], softSurfer License [softsurfer], SolderPad Hardware License v0.5 [shl-0.5], Solderpad Hardware License v2.0 [shl-2.0], Solderpad Hardware License v2.1 [shl-2.1], SolderPad Hardware License, Version 0.51 [shl-0.51], Sparky License [sparky], SpeechWorks Public License 1.1 [speechworks-1.1], SQLite Blessing [blessing], Standard ML of New Jersey [standard-ml-nj], Stanford PVRG License [stanford-pvrg], STLport License 2000 [stlport-2000], STLport License 4.5 [stlport-4.5], STREAM Benchmark License [stream-benchmark], Stu Nicholls License [stu-nicholls], Sun RPC License [sun-rpc], Sun source code License [sun-source], SunPro Attribution License [sunpro], Sunsoft License [sunsoft], Supervisor License [supervisor], svndiff License [svndiff], SWIG Library License [swig], Symlinks License [symlinks], Symphonysoft [symphonysoft], Synopsys MIT License [synopsys-mit], Synthesis Toolkit License [synthesis-toolkit], SystemC Open Source License Agreement [accellera-systemc], Taiwan Open Government Data License, version 1.0 [ogdl-taiwan-1.0], Takao Abe License [takao-abe], Takuya OOURA License [takuya-ooura], Talis Community License [ttcl], Tatu Ylonen License [tatu-ylonen], TCG Spec License v1 [tcg-spec-license-v1], TCL/TK License [tcl], TCP Wrappers License [tcp-wrappers], TekHVC License [tekhvc], Term Readkey License [term-readkey], Tested Software License [tested-software], TeX Live License [tex-live], Text-Tabs+Wrap License [ttwl], TFL [tfl], The Happy Bunny License [happy-bunny], Theodore Ts'o license [tso-license], Things I Made (TIM) Public License [things-i-made-public-license], Tidy License [tidy], Tiger Cryptography License [tiger-crypto], Tigra Calendar 3.2 License [tigra-calendar-3.2], Tigra Calendar 4.0 License [tigra-calendar-4.0], Tim Janik License 2003 [tim-janik-2003], Time::ParseDate License [tpdl], Timestamp Picker License [timestamp-picker], TTYP0 License [ttyp0], TU Berlin License 1.0 [tu-berlin], TU Berlin License 2.0 [tu-berlin-2.0], Tumbolia Public License [tumbolia], TwistedSNMP License [twisted-snmp], UCAR License [ucar], UnboundID LDAP SDK Free Use License [ldap-sdk-free-use], Unicode DFS 2015 [unicode-dfs-2015], Unicode DFS 2016 [unicode-dfs-2016], Unicode Inc License Agreement [unicode], Unicode Mappings License [unicode-mappings], University of British Columbia License [ubc], University of Michigan OSL [michigan-disclaimer], UNIX Network Programming Book License [unpbook], UnixCrypt License [unixcrypt], Unlicense [unlicense], Unlimited Binary Use Exception [unlimited-binary-use-exception], UPL 1.0 [upl-1.0], US Government Public Domain [us-govt-public-domain], US Government Unlimited Rights [us-govt-unlimited-rights], USRobotics Permissive License [usrobotics-permissive], Utopia Typeface License [utopia], VCalendar License [vcalendar], Vic Metcalfe Public Domain [vic-metcalfe-pd], VIM License [vim], Visual Idiot [visual-idiot], Visual Numerics License [visual-numerics], Vixie Cron License [vixie-cron], Vovida Software License 1.0 [vsl-1.0], W3C 3-Clause BSD License [w3c-03-bsd-license], W3C Software Notice and License [w3c], W3C-SOFTWARE-19980720 [w3c-software-19980720], W3C-SOFTWARE-DOC-20150513 [w3c-software-doc-20150513], w3m License [w3m], Westhawk License [westhawk], Whistle Communications License [whistle], Whitecat License [whitecat], WIDE License [wide-license], Wide Open License [wol], Widget Workshop License [widget-workshop], William Alexander License [william-alexander], wingo License [wingo], Wordnet License [wordnet], Wrox Press License [wrox], WS-Addressing Specification License [ws-addressing-spec], WS-Policy Specification [ws-policy-specification], WS-Trust Specification [ws-trust-specification], Wsuipa License [wsuipa], WTFNMFPL-1.0 [wtfnmfpl-1.0], WTFPL 1.0 [wtfpl-1.0], WTFPL 2.0 [wtfpl-2.0], WTHPL 1.0 [wthpl-1.0], wxWidgets Licence [wxwidgets], wxWindows Unrestricted Licence 3.0 [wxwindows-u-3.0], X11 Documentation License [x11-doc], X11 License [x11], X11-R5 [x11-x11r5], X11-Style (Acer) [x11-acer], X11-Style (Adobe) [x11-adobe], X11-Style (Adobe-DEC) [x11-adobe-dec], X11-Style (Bitstream Charter) [x11-bitstream], X11-Style (David R. Hanson) [x11-hanson], X11-Style (DEC 1) [x11-dec1], X11-Style (DEC 2) [x11-dec2], X11-Style (DSC Technologies) [x11-dsc], X11-Style (FSF) [x11-fsf], X11-Style (Keith Packard) [x11-keith-packard], X11-Style (Lucent) [x11-lucent], X11-Style (Lucent-variant) [x11-lucent-variant], X11-Style (OAR) [x11-oar], X11-Style (Open Group) [x11-opengroup], X11-Style (OpenGL) [x11-opengl], X11-Style (Quarterdeck) [x11-quarterdeck], X11-Style (Realmode) [x11-realmode], X11-Style (Silicon Graphics) [x11-sg], X11-Style (Stanford University) [x11-stanford], X11-Style (Tektronix) [x11-tektronix], X11-Style (Tiff) [x11-tiff], X11-Style (X Consortium Veillard) [x11-xconsortium-veillard], X11-Style (X Consortium) [x11-xconsortium], Xdebug License v 1.03 [xdebug-1.03], XFree86 License 1.0 [xfree86-1.0], XFree86 License 1.1 [xfree86-1.1], xinetd License [xinetd], XML:DB Initiative Software License 1.0 [xmldb-1.0], XSkat License [xskat], xxd License [xxd], Yale CAS License [yale-cas], Yensdesign License [yensdesign], Zed License [zed], Zend Engine License 2.0 [zend-2.0], ZeusBench notice [zeusbench], ZLIB License [zlib], ZLIB License with Acknowledgment [zlib-acknowledgement], ZPL 1.0 [zpl-1.0], ZPL 1.1 [zpl-1.1], ZPL 2.0 [zpl-2.0], ZPL 2.1 [zpl-2.1], zsh License [zsh], Zuora Software License [zuora-software], Zveno Research License [zveno-research]
\end{simplechar}

The list above gives the short name (or name, if no short name exists) along with the key, in square brackets, from the ScanCode license dataset available at \url{https://github.com/aboutcode-org/scancode-toolkit/tree/develop/src/licensedcode/data/licenses}.

\end{document}